\documentclass[a4paper]{article}

\usepackage{graphicx}

\usepackage{authblk}
\usepackage{amsmath}
\usepackage{amsthm}
\usepackage{amssymb}
\usepackage{booktabs}
\usepackage{xspace}
\usepackage[margin=1.25in]{geometry}
\usepackage{hyperref}

\newcommand{\bs}{\mathbf{s}}
\newcommand{\bq}{\mathbf{q}}
\newcommand{\br}{\mathbf{r}}
\newcommand{\bx}{\mathbf{x}}

\newcommand{\be}{\mathbf{e}}
\newcommand{\ba}{\mathbf{a}}
\newcommand{\bb}{\mathbf{b}}

\newcommand{\bt}{\mathbf{t}}

\newcommand{\bzero}{\mathbf{0}}

\newcommand{\bV}{\mathbf{V}}
\newcommand{\bI}{\mathbf{I}}

\newcommand{\bT}{\mathbf{T}}
\newcommand{\bp}{\mathbf{p}}

\newcommand{\free}{free\xspace}
\newcommand{\Free}{Free\xspace}
\newcommand{\bound}{causal\xspace}
\newcommand{\Bound}{Causal\xspace}

\newcommand{\piv}{\mathrm{piv}}
\newcommand{\supp}{\mathrm{supp}}
\newcommand{\deff}{:\ }

\newtheoremstyle{TheoremDefStyle}
{\topsep}   
{\topsep}   
{\itshape}  
{30pt}       
{\bfseries} 
{:}         
{5pt plus 1pt minus 1pt} 
{}          

\theoremstyle{TheoremDefStyle}

\numberwithin{equation}{section}

\title{State Algebra for Propositional Logic}

\author[1]{Dmitry Lesnik}
\author[1,2]{Tobias Sch{\"a}fer}
\affil[1]{Stratyfy Inc., New York, New York, USA}
\affil[2]{Department of Mathematics, College of Staten Island, Staten Island, NY, USA \& Physics Program, CUNY Graduate Center, NY, USA}

\date{}

\begin{document}

\maketitle

\begin{abstract}
    This paper presents State Algebra, a novel framework designed to represent and manipulate propositional logic using algebraic methods. The framework is structured as a hierarchy of three representations: Set, Coordinate, and Row Decomposition. These representations anchor the system in well-known semantics while facilitating the computation using a powerful algebraic engine. A key aspect of State Algebra is its flexibility in representation. We show that although the default reduction of a state vector is not canonical, a unique canonical form can be obtained by applying a fixed variable order during the reduction process. This highlights a trade-off: by foregoing guaranteed canonicity, the framework gains increased flexibility, potentially leading to more compact representations of certain classes of problems. We explore how this framework provides tools to articulate both search-based and knowledge compilation algorithms and discuss its natural extension to probabilistic logic and Weighted Model Counting.
\end{abstract}

\newpage

\tableofcontents

\newpage

\section{Introduction}\label{sec:introduction}

This study introduces \emph{State Algebra}, a generic framework for representing and manipulating propositional logic by casting it into an algebraic form. State Algebra considers a system of Boolean variables and the corresponding space of all possible truth assignments. Each assignment represents a \emph{state}, and a logical formula is identified with a subset of states, a \emph{state vector}, in which it holds true. This set-theoretic approach reduces logical inference to algebraic operations on the state vectors.

From an algorithmic perspective, State Algebra presents a compact data structure suitable for performing calculations in propositional logic. It replaces the symbolic language of logical formulas with sparse matrices populated with zeros, ones, and missing values (holes) and algebraic operations on them. Similar to how Reduced Ordered Binary Decision Diagrams (ROBDD)~\cite{Bryant1992} offer a compact graph representation of Boolean functions, State Algebra introduces a compact algebraic representation that can be manipulated using familiar notations, such as addition and multiplication.

One important distinction from state-of-the-art data structures, most notably ROBDDs, is the concept of canonicity. A canonical representation of the ROBDD guarantees a unique form for any given Boolean function, a property that is the cornerstone of its efficiency for equivalence-heavy applications, such as formal verification. State Algebra is flexible and supports both canonical and noncanonical representations. By fixing the variable order and systematically applying the reduction rules, a canonical form can be achieved. However, a noncanonical approach, which is more flexible and not tied to strict variable ordering, may represent certain functions more compactly. It is plausible that for some classes of problems, a ``good enough'' compact representation in State Algebra is easier to find than an optimal global variable ordering. The freedom to sacrifice canonicity in exchange for greater representational flexibility is a characteristic feature of the state-algebraic approach.

State Algebra introduces three representations of the state space as a hierarchy of abstractions: semantic (Set), algebraic (Coordinate), and computational (Row Decomposition). This progression allows the framework to be grounded in the familiar semantics of set theory while leveraging a powerful algebraic engine for the computation.

\emph{Set Representation} serves as the semantic foundation, establishing a link to the model-theoretic view of logic: the meaning of a formula is the set of all models in which it is true.

\emph{Coordinate Representation} generalises this representation by viewing the state space as a formal algebraic structure, transforming logical manipulation into algebraic computation. Based on this, the formalism of ``t-objects'' provides a framework for a purely abstract-algebraic viewpoint on propositional logic.

\emph{Row Decomposition} is the concrete data structure for implementation, representing state vectors as a sum of t-objects.

State Algebra can be viewed as a language for representing logical functions and a set of algebraic tools for manipulating them. While it is not an optimisation algorithm in itself, it provides machinery to express the principles behind modern solvers. An implementer can choose between a knowledge compilation approach (building a complete representation of the solution space, similar to BDDs) and a search-based approach (which is typical for SAT solvers). Optimisation heuristics, such as Davis-Putnam-Logemann-Loveland (DPLL)~\cite{Prasad2005,Malik2009} and Conflict-Driven Clause Learning (CDCL)~\cite{Marques1999,Moskewicz2001}, could be adapted to guide manipulations within the State Algebra framework.

A powerful feature of this formalism is its natural extensibility to probabilistic logic (aka Markov Random Fields) and higher-order logic. The extension to probabilistic logic, in particular, enables efficient algorithms for Weighted Model Counting (WMC). Both topics deserve separate discussions and are beyond the scope of this paper, but are outlined in the section on future work.

The remainder of this paper is organised as follows. Section ``Terminology'' provides a quick reference for the key concepts introduced throughout this paper. In Section~\ref{sec:basic-definitions}, we introduce the basic definitions in State Algebra, such as the state space and reduced matrix notation. Section~\ref{sec:propositional-logic} establishes a link between symbolic logic and state-algebraic representation. A more formal definition of state-algebraic concepts is considered in Section~\ref{sec:algebra-of-state-vectors}, where we introduce the hierarchy of representations: Set, Coordinate, and Row Decomposition. Section~\ref{sec:algebra-of-t-objects} is dedicated to the abstract formalism of t-objects. We consider some of the most common implementation algorithms and assess their complexities in Section~\ref{sec:algorithms}. As mentioned earlier, this study does not focus on optimisation heuristics. The algorithms are described as proof of concept, and the estimated complexity serves as an upper-bound assessment. In Section~\ref{sec:enhancements-and-future-work}, we briefly outline how the proposed state-algebraic formalism can be extended to probabilistic and higher order logic.

\section{Terminology}\label{sec:terminology}

In this section, we briefly introduce the main concepts which will be defined more formally in the main text. This serves as a glossary for quick reference purposes.

\begin{itemize}
    \item \textbf{State space} $S_\mathcal{E}$ is defined as the collection of all possible truth assignments for a given set of Boolean variables. Each individual truth assignment is referred to as a \textbf{state}. A \textbf{state vector} is a subset of state space. We consider three different ways of interpreting state vectors: \emph{set representation}, \emph{coordinate representation}, and \emph{row decomposition}.

    \item \textbf{Matrix notation}. This is a way to display a state vector as a matrix, for example,
    \[
        \bs = \begin{Bmatrix}
                  1 & - \\
                  - & 0
        \end{Bmatrix}
    \]
    In matrix notation, the dashes denote \textbf{holes}. A row containing holes signifies a set of rows in which all possible combinations of 0s and 1s are substituted for holes.

    \item \textbf{T-object} is a single row of a state vector in matrix notation. If a t-object contains holes, it represents a multitude of states corresponding to all possible assignments of the truth values to the holes.

    \item \textbf{Set representation} interprets a state vector as a set of states, where duplicates are disregarded. The space of the state vectors in this representation, $\mathbb{S}^*_\mathcal{E}$, corresponds to the power set of $S_\mathcal{E}$. This representation allows the use of set operations such as $\cup$, $\cap$, and $\smallsetminus$.

    \item \textbf{Coordinate representation} is a way to interpret a state vector with states having integer multiplicity factors (coordinates). With coordinate-wise operations of addition, subtraction, multiplication, and scaling, the space of state vectors has a structure of $\mathbb{Z}$-algebra. We designate this space as $\mathbb{S}_\mathcal{E}$. The relationship between $\mathbb{S}_\mathcal{E}$ and $\mathbb{S}^*_\mathcal{E}$ is established by the following facts:
    \begin{itemize}
        \item Every element of $\mathbb{S}^*_\mathcal{E}$ is an equivalence class within $\mathbb{S}_\mathcal{E}$.
        \item Binary vectors in $\mathbb{S}_\mathcal{E}$ (i.e., those with only 0 or 1 coordinates) have one-to-one mapping to $\mathbb{S}^*_\mathcal{E}$.
    \end{itemize}

    \item \textbf{Identity transformation} is a modification in matrix notation with the help of holes, which keeps the coordinate representation unchanged. For instance:
    \[
        \begin{Bmatrix}
            1 & 1 \\
            1 & 0 \\
            0 & 0
        \end{Bmatrix} =
        \begin{Bmatrix}
            1 & - \\
            0 & 0
        \end{Bmatrix} =
        \begin{Bmatrix}
            1 & 1 \\
            - & 0
        \end{Bmatrix}
    \]

    \item \textbf{Equivalence transformation} is a transformation which changes coordinates within the equivalence class. The equivalence transformation keeps the set representation unchanged but can change the multiplicity factors of the individual states. For instance:
    \[
        \begin{Bmatrix}
            1 & - \\
            - & 1
        \end{Bmatrix} \simeq
        \begin{Bmatrix}
            1 & 0 \\
            - & 1
        \end{Bmatrix}
    \]

    \item \textbf{Row decomposition} interprets a state vector as a sum of its rows -- t-objects. Row decomposition is a subset of $\mathbb{S}_\mathcal{E}$ with only nonnegative coordinates. It supports coordinate-wise addition and multiplication operations, although neither operation has an inverse. We designate the space of the state vectors obtained by row decomposition as $\mathbb{S}^+_\mathcal{E}$. For the elements of $\mathbb{S}^+_\mathcal{E}$, instead of subtraction, we introduce an operation $\smallsetminus$, which is similar to the set subtraction.
\end{itemize}

\section{Basic Definitions}\label{sec:basic-definitions}

In this section, we introduce some basic definitions of State Algebra, such as state space and state vector, as well as reduced matrix notation. An elementary building block of the state space is an event, a Boolean variable that can be assigned a truth value of 1 (true) or 0 (false). In propositional logic, events are simple objects without any internal structure. We will use the names event and variable interchangeably.

\subsection{State Space}\label{subsec:state-space}

An event $E$ is a Boolean variable that can take values 0 and 1. Let $\mathcal{E} = \{E_1,\ldots,E_N\}$ be a finite system of events comprising $N$ elements. Consider one assignment of truth values to the events, and let $e_i$ be the value of $i$-th event. A vector of values
\[
    s = \{ e_1,\ldots,e_N\}\,,\qquad e_i\in\{1,0\}
\]
represents a \textbf{state} in $\mathcal{E}$. A \textbf{state space} $S_\mathcal{E}$ denotes the set of all possible states. For a finite~$N$, the state space $S_\mathcal{E}$ contains $2^N$ states.

\subsubsection{State Vector in Matrix Notation}

An arbitrary set of states is referred to as a \textbf{state vector}. A state vector can be displayed as a matrix, the rows of which are interpreted as states:
\[
    \bs =
    \begin{Bmatrix}
        e_{11} & e_{12} & \ldots & e_{1N} \\
        e_{21} & e_{22} & \ldots & e_{2N}  \\
        \cdots & & & \\
        e_{m1} & e_{m2} & \ldots & e_{mN}
    \end{Bmatrix} \,.
\]
We refer to this method of displaying a state vector as \textbf{matrix notation}. We will use two different ways of interpreting the state vectors in matrix notation. Interpretation, in which we treat a state vector as a set of states (and hence ignore possible state duplicates), is called a \textbf{set representation}. The state vectors in the set representation form a space of all possible subsets of $S_\mathcal{E}$. We designate this space as $\mathbb{S}^*_\mathcal{E}$.
\begin{align}
    \mathbb{S}^*_\mathcal{E} = \mathcal{P}(S_\mathcal{E})
\end{align}
where $\mathcal{P}()$ denotes the power set. Formally, the set representation is an instance of Boolean algebra, equipped with set operations of union, intersection, and complement.

Below, we introduce a more generic \emph{coordinate representation} which accounts for possible state duplicates (see Sec.~\ref{subsec:coordinate-representation}). In our discussion of inference algorithms, we also introduce \emph{row decomposition}, a truncated version of the coordinate representation that uses only nonnegative coordinates. A state vector in row decomposition is represented as the sum of its rows (see Sec.~\ref{subsec:row-decomposition}).

We use bold Latin letters
\[\bp, \bq, \br, \bs, \bx, \bV, \bI\]
to denote the state vectors. The number $m$ of distinct states in a state vector is called \textbf{cardinality}. Obviously, the cardinality satisfies $0\leq m \leq 2^N$. If $m=0$ we say that the state vector is \textbf{empty}, and designate it as
\[
    \bzero
\]

If $m = 2^N$ we say that the state vector is \textbf{trivial} and designate it as
\[
    \bt
\]

Let $\mathcal{K}$ be a subset of the system of events
\[
    \mathcal{K} = \{E_{i_1}, \ldots, E_{i_m}\} \subset \mathcal{E}
\]
The state vector obtained by selecting columns $i_1,\ldots,i_m$ of the original state vector is called a state vector \emph{constrained} to $\mathcal{K}$.

An event is said to be \textbf{identically true} in a vector $\bs$ if it has a value of 1 in all the states of the state vector. Similarly, an event is said to be \textbf{identically false} in $\bs$ if it has a value of zero in all states of~$\bs$. If a column contains both values, 0 and 1, the corresponding event is said to have an indefinite value.

The purpose of logical inference can often be reduced to proving that a particular event is true or false given a premise (a set of logical rules). In terms of State Algebra, this is equivalent to showing that the target event is identically true or false in the allowed state space, that is, in the subset of state space satisfying all logical rules of the premise.

\subsubsection{\Bound and \Free Events}

We say that an event $E_i$ is \textbf{\free} in a state vector $\bs$ if for every state $s$ which is an element of $\bs$ a state $s'$ obtained by inverting the value of $E_i$ is also an element of $\bs$:
\begin{align*}
    &\forall  s = \{e_1, \ldots, e_{i-1}, e_i, e_{i+1}, \ldots\} :\\
    & s' = \{e_1, \ldots, e_{i-1}, 1-e_i, e_{i+1}, \ldots\}, \\
    & s \in \bs \quad \iff \quad s' \in \bs\,,
\end{align*}
where $e_i$ is the value of the event $E_i$.

Events that are not \free in the state vector $\bs$ are called \textbf{\bound}. We will refer to the set of \bound events as \textbf{support} and use notation $\supp(\bs)$, for example,
\[
    \supp(\bs) = \{i, j, k\}
\]
implies that $\{E_i, E_j, E_k\}$ are \bound events of $\bs$.

In the following example of a state vector on the space $\{E_1, E_2, E_3\}$
\[
    \bs =
    \begin{Bmatrix}
        0 & 1 & 0\\
        0 & 1 & 1\\
        1 & 0 & 0\\
        1 & 0 & 1\\
    \end{Bmatrix}\,,
\]
the \bound events are $E_1$ and $E_2$, whereas event $E_3$ is \free. Hence
\[
    \supp(\bs) = \{1, 2\}
\]

\subsection{Reduced Representation}\label{subsec:reduced-representation}

In propositional logic, a state vector containing states allowed by a logical rule (or a set of rules) is known as a truth table. The latter is usually helpful for educational and illustration purposes, but is hardly usable in practice because the exponentially growing size makes numerical methods involving truth tables prohibitively expensive. In State Algebra, a reduced representation of state vectors allows the encoding of the truth table in a compact way, making it a powerful vehicle for numeric algorithms.

\subsubsection{Hole Notation}

Let us consider a state vector $\bs$ in which some event (let it for example be the first event $E_1$) has an undefined value, and other events have fixed values. The state vector contains two states:
\[
    \bs =
    \begin{Bmatrix}
        0 & e_2 & \ldots & e_N \\
        1 & e_2 & \ldots & e_N
    \end{Bmatrix} \,.
\]

We introduce a ``\textbf{hole}'' notation for an undefined event value and designate it as a dash ``--''. By definition, if a state in the state vector has a hole at the $i$-th position, then it is equivalent to two states having 1 and 0 in the $i$-th position and identical otherwise. For instance, the latter state vector can be represented as
\begin{align}
    \bs =
    \begin{Bmatrix}
        0 & e_2 & e_3 & \ldots & e_N \\
        1 & e_2 & e_3 & \ldots & e_N
    \end{Bmatrix} =
    \begin{Bmatrix}
        - & e_2 & e_3 & \ldots & e_N
    \end{Bmatrix} \,.
\end{align}
Here is another example of a state vector consisting of three states, where two states can be represented as one row containing a hole:
\[
    \begin{Bmatrix}
        1 & 1 & 1  \\
        1 & 0 & 1  \\
        0 & 0 & 0
    \end{Bmatrix} =
    \begin{Bmatrix}
        1 & - & 1  \\
        0 & 0 & 0
    \end{Bmatrix}.
\]

This notation is straightforward to generalise for an arbitrary number of holes in a row. If a row in a state vector contains $n$ holes, it is equivalent to $2^n$ different states with all possible assignments of 0s and 1s to the holes. For example:
\[
    \begin{Bmatrix}
        - & - & e_3 & \ldots & e_N
    \end{Bmatrix} =
    \begin{Bmatrix}
        0&0& e_3&\ldots  & e_N\\
        0&1& e_3&\ldots  & e_N\\
        1&0& e_3&\ldots  & e_N\\
        1&1& e_3&\ldots  & e_N
    \end{Bmatrix} \,.
\]

Notice that with the introduced notation, the trivial state vector can be written in the form
\[
    \bt =
    \begin{Bmatrix}
        - & - & \ldots  & -
    \end{Bmatrix} \,.
\]

An \textbf{atomic reduction} is a transformation that replaces two rows of a state vector with a single equivalent row using the ``hole'' notation. A pair of rows is \textbf{reducible} if an atomic reduction can be applied to them. If we want to specify the index of an event whose value was replaced by a hole, we refer to it as an \emph{atomic reduction by index} $i$ or \emph{by variable} $E_i$. Here is an example of an atomic reduction by index 2 applied to the first two rows:
\[
    \begin{Bmatrix}
        1 & 1 & -\\
        1 & 0 & - \\
        0 & 0 & 1
    \end{Bmatrix} = \begin{Bmatrix}
                        1 & - & -\\
                        0 & 0 & 1
    \end{Bmatrix}
\]
The inverse operation of atomic reduction is called \textbf{atomic expansion} by index $i$.

For any state vector, atomic reduction can be applied to any reducible pair of rows. This operation can be performed recursively until no more reducible rows exist. We define two transformations for the state vectors: \textbf{expansion} and \textbf{reduction}. Reduction is the application of atomic reduction once or multiple times, whereas expansion is the reverse transformation. A state vector is \textbf{expanded} if it contains no holes and \textbf{reduced} (or \textbf{irreducible}) if no rows allow atomic reduction.

Expansion and reduction only affect the matrix representation of a state vector, not the actual set of states it comprises. The cardinality (number of unique states) of a state vector remains constant, regardless of whether it is in an expanded, reduced, or partially reduced form. We refer to expansion and reduction as \textbf{identity transformations}, and two state vectors are considered \textbf{identical} if one can be obtained from the other through an identity transformation.

\subsubsection{Canonical Reduction}

The reduced form of the state vector is generally nonunique. The reduction result depends on the order in which the atomic reduction is applied. For instance, the following state vectors which represent the function $(E_1 \vee E_2)$ are identical:
\[
    \begin{Bmatrix}
        1 & 1 \\
        1 & 0 \\
        0 & 1
    \end{Bmatrix} = \begin{Bmatrix}
                        1 & - \\
                        0 & 1
    \end{Bmatrix} = \begin{Bmatrix}
                        - & 1 \\
                        1 & 0
    \end{Bmatrix}
\]
A stronger statement can be made: different reduced forms of a state vector can have different sizes (i.e. number of rows). For example, a trivial state vector (representing the constant $True$ function) for $N=3$ has an optimal representation of a single row $\{-\ -\ -\}$. However, the following is a valid irreducible representation of the same function as well:
\begin{align}
    \label{eq:trivial_irreducible}
    \bt = \begin{Bmatrix}
              0 & 0 & 0 \\
              1 & 1 & 1 \\
              - & 0 & 1 \\
              1 & - & 0 \\
              0 & 1 & -
    \end{Bmatrix} =
    \begin{Bmatrix}
        - & - & -
    \end{Bmatrix}
\end{align}

A unique or \textbf{canonical} reduced form can be achieved by imposing a fixed order on the application of atomic reduction, according to the following steps:
\begin{enumerate}
    \item Fix an order of events: $E_1 < E_2 < \ldots < E_N$.
    \item Apply atomic reduction exhaustively for all reducible pairs of rows by index $i=N$, then by index $i=N-1$, and so on, down to $i=1$.
\end{enumerate}
The uniqueness of the ordered reduction follows from the fact that at each step, the set of reducible pairs of rows with respect to event $E_i$ is uniquely determined. This produces a unique input for the next reduction step on event $E_{i-1}$, thus making the entire process deterministic by induction.

Expressed in the language of Term Rewriting Systems (TRS), the ordered reduction procedure makes State Algebra a canonical TRS, which guarantees that each term (state vector) can be reduced to a unique \emph{normal form}.

Canonical reduction offers the advantage of uniqueness, making the equivalence check of the two state vectors a trivial identity check. However, the degree of state vector compression in the canonical form is highly dependent on the selected variable ordering. In contrast, noncanonical reduction can employ various heuristics to potentially achieve greater compression and enhance the efficiency of certain inference algorithms. Therefore, the decision between canonical and noncanonical reductions should be made based on the specific problem at hand.

\subsubsection{State Duplicates}

It is possible that a state vector in matrix notation in a (fully or partially) reduced form has some states that are repeated in different rows, although all the rows are distinct. For instance, in the following vector
\[
    \begin{Bmatrix}
        - & 1 \\
        0 & -
    \end{Bmatrix}\,,
\]
state $\{0\ 1\}$ is present in both rows. If interpreted as a set of states, the state duplicates in the state vector are ignored.

\subsubsection{Trivial Columns}

For an arbitrary state vector $\bs$ (regardless of whether it is in a fully or partially reduced form), we call those columns \textbf{trivial}, which contain only holes. For instance, in the following state vector defined on $\{E_1, E_2, E_3\}$
\[
    \bs = \begin{Bmatrix}
              0 & - & - \\
              1 & 0 & -
    \end{Bmatrix}
\]
the last column is trivial.

With this definition, the following statements are equivalent:
\begin{itemize}
    \item An event $E_i$ is \free in the state vector $\bs$.
    \item There exists a reduced form of $\bs$ in which the column $i$ is trivial.
\end{itemize}
A trivial column in a state vector is always \free. However, the opposite is not always true. Different statements can be made depending on whether canonical or noncanonical reduction is used.

In a canonically reduced state vector without duplicates, the column corresponding to a free variable is always trivial.
\begin{proof}
    Let $E_i$ be a free variable for the function represented by the state vector $\bs$. This means that for every row in $\bs$, there exists a ``twin'' row that is identical in all columns except for column $i$, where it has the opposite value. Let us partition the rows of $\bs$ into two sets, $\bs^+$ and $\bs^-$, based on their values in column $i$. Apart from column $i$ the sets $\bs^+$ and $\bs^-$ are identical.

    The ordered reduction algorithm processes variables from $E_N$ down to $E_1$. For any variable $E_j$ with $j>i$, atomic reductions can only occur within $\bs^+$ or within $\bs^-$, but never between a row from $\bs^+$ and a row from $\bs^-$ because their values in column $i$ differ.

    By the time the algorithm is ready to perform reductions on column $i$, the sets $\bs^+$ and $\bs^-$ will have been independently reduced, and the resulting reduced sets of rows will be identical in every column, except for $i$. This creates a perfect pairing for atomic reduction on variable $i$ for each row. Consequently, the entire $i$-th column is replaced with ``hole'' entries, making it a trivial column.
\end{proof}

If not in a canonical form, or if state duplicates are present, irreducible forms of a state vector may exist, in which some \free columns are not trivial. For example, all the columns of a trivial vector $\bt$ are \free. However, the irreducible representation of $\bt$ in Eq.~\eqref{eq:trivial_irreducible} has three nontrivial columns. An algebraic method exists for determining whether an event is \free in a state vector (see Appendix~\ref{sec:other-operations-on-state-vectors}).

Columns which are not trivial are called \textbf{pivot} columns and are designated $\piv(\bs)$. Evidently, the set of pivot columns is a superset of the \bound columns.
\[
    \supp(\bs) \subseteq \piv(\bs)
\]
In the following example of a state vector
\[
    \bs = \begin{Bmatrix}
              1 & - & - \\
              1 & 0 & -
    \end{Bmatrix}
\]
only the first column is \bound. Hence, $\supp(\bs) = \{1\}$ but $\piv(\bs) = \{1, 2\}$.

Pivot columns are a useful concept for optimising algebraic transformation algorithms. Some optimisation heuristics use \bound columns of the state vector. However, identifying the \bound columns of a state vector may be computationally expensive. Pivot columns serve as a reasonable and computationally efficient approximation of \bound columns.

\subsubsection{Orthogonality and Independence}

Here, we introduce two useful definitions. We call these two state vectors \textbf{independent} and designate them as $\bs \wr \bq$ if their supports do not overlap.
\begin{align*}
    \bs \wr \bq\qquad \iff \qquad \supp(\bs)\cap\supp(\bq) = \varnothing
\end{align*}
For example, the following two state vectors are independent:
\begin{align*}
    \bs = \begin{Bmatrix}
              1 & 1 & - & - \\
              0 & - & - & -
    \end{Bmatrix}\,,\quad
    \bq = \begin{Bmatrix}
              - & - & 0 & 0 \\
              - & - & 1 & -
    \end{Bmatrix}\quad \Rightarrow \quad \bs\wr \bq
\end{align*}

We call two state vectors \textbf{orthogonal} and designate them as $\bs \perp \bq$ if their intersection is empty:
\begin{align*}
    \bs\perp\bq \qquad \iff \qquad \bs \cap \bq = \varnothing
\end{align*}
For example, the following two vectors are orthogonal:
\begin{align*}
    \bs = \begin{Bmatrix}
              0 & 1 & 1 & - \\
              1 & 0 & 0 & -
    \end{Bmatrix}\,,\quad
    \bq = \begin{Bmatrix}
              - & 1 & 0 & 0 \\
              - & 0 & 1 & 0
    \end{Bmatrix}\quad \Rightarrow \quad \bs \perp \bq
\end{align*}
Note that it is not possible for two vectors to be independent and orthogonal simultaneously. While this statement is not difficult to prove, it is a great exercise for interested readers.

\section{Propositional Logic}\label{sec:propositional-logic}

A logical formula is a Boolean function defined on states
\[
    f: S_\mathcal{E} \to B
\]
where $B = \{1, 0\}$ is a Boolean set and $S_\mathcal{E} = B^N$ is the state space. We say that the state $s$ satisfies formula $f$ if $f(s) = 1$ and falsifies the formula if $f(s) = 0$. Each logical formula divides the set of all possible states into two complementary subsets: the states satisfying the formula (we will sometimes refer to these states as \emph{allowed} states or \emph{valid} states) and those falsifying the formula (\emph{prohibited} states).

\subsection{Logical Formulas In State Algebra}\label{subsec:logical-formulas-in-state-algebra}

For every logical formula $f$, we can construct a state vector $\bs$ that consists of all states that satisfy~$f$. We use the following notation
\[
    f\sim \bs\qquad \iff \qquad\bs = \{s: f(s) = 1\}
\]
We say that the formula $f$ has a corresponding state vector $\bs$. We refer to the state vectors containing the allowed states of logical formulas as \textbf{valid} state vectors. This is to distinguish them from the information state vectors (see Sec.~\ref{subsubsec:information_s_vectors}). If not specified explicitly, we will always imply that a state vector is a valid state vector.

A formula is called a \emph{tautology} if its state vector is trivial, and a \emph{contradiction} if its state vector is empty.

Let us consider, for example, the logical formula $f\deff (E_1\,\to\,E_2)$. The state vector corresponding to this formula has in the space $\{E_1, E_2\}$ the following three states:
\[
    E_1\,\to\,E_2\quad \sim\quad
    \bs =
    \begin{Bmatrix}
        1 & 1 \\
        0 & 1  \\
        0 & 0
    \end{Bmatrix}
\]

The state $\{1\  0\}$ is not in $\bs$, as the formula $f$ is falsified if $E_1$ is true and $E_2$ is false.

We define the support of a formula as the support of the corresponding state vector:
\[
    \supp(f) = \supp(\bs)
\]
where $f \sim\bs$.

The reduced representation of the state vectors provides a method for efficiently representing logical formulas in terms of the corresponding state vectors. This representation remains compact for arbitrary numbers of variables because, for every logical formula, only its \bound columns need explicit truth values. For instance, a state vector representing implication formula $E_1 \to E_2$ can be written as
\[
    E_1 \to E_2 \quad \sim \quad
    \begin{Bmatrix}
        1 & 1 & - & \ldots & - \\
        0 & - & - & \ldots & -
    \end{Bmatrix}
\]
As another example, for disjunction, we obtain
\[
    E_1\, \vee \, E_2 \qquad \sim \qquad
    \begin{Bmatrix}
        1 & -  & - & \ldots & -\\
        0 & 1  & - & \ldots & -
    \end{Bmatrix}
\]

Obtaining a compact state vector representation of an arbitrary compound logical formula is straightforward. The brute-force algorithm includes the following steps:
\begin{itemize}
    \item For the \bound columns of the logical formula, construct a truth table corresponding to the allowed states of the logical formula.
    \item Optionally, reduce the obtained truth table.
    \item Complete the state vector by adding trivial columns for all remaining variables.
\end{itemize}
It is often practical to use supplementary events to construct the state vector of compound logical formulas.

\subsubsection{Supplementary Events}\label{subsubsec:suppl_ev}

To determine the state vector for a compound logical formula, it is helpful to break it down into simpler formulas by introducing supplementary events. For example, consider the following formula:
\[
    E_1 = (E_2\, \wedge\,(E_3 \, \to \, E_4)).
\]
We can introduce a supplementary event, $E_s$, for the inner brackets, transforming the compound formula into a conjunction of simpler formulas, as follows:
\begin{align*}
    & E_s = (E_3 \, \to \, E_4) \\
    & E_1 = (E_2\, \wedge \,E_s)
\end{align*}
Once the state vectors are established for each simpler formula, their intersection (viewed as sets of states) yields the state vector of the original compound formula. The supplementary events can subsequently be removed from the system, as detailed in Appendix~\ref{subsubsec:removing-supplementary-events}.

\subsubsection{Extended Formulas}

Consider the logical formula $f$. Since the value of $f$ is boolean, we can introduce a new event, $E_k$, and define an extension of $f$ as
\[
    F \deff\; (E_k = f)\,.
\]
We refer to $E_k$ as an \emph{indicator event} for $f$.

For example, consider the formula $E_2 \rightarrow E_3$. Its extension with the indicator event $E_1$ is $E_1 = (E_2 \rightarrow E_3)$. The corresponding state vector in space ${E_1, E_2, E_3}$ is
\[
    F \quad \sim\quad \bs =
    \begin{Bmatrix}
        1 & 1 & 1\\
        1 & 0 & 1\\
        1 & 0 & 0\\
        0 & 1 & 0
    \end{Bmatrix} =
    \begin{Bmatrix}
        1 & 1 & 1\\
        1 & 0 & -\\
        0 & 1 & 0
    \end{Bmatrix}
\]
If $f$ is a tautology, then its indicator event $E_k$ in the state vector of $F$ is identically true. If $f$ is a contradiction, $E_k$ is identically false.

Note that the extended formulas are valid on the entire state space which does not include supplementary events, and hence can always be added to the premise. They are particularly valuable for solving a specific type of logical inference problem: determining the validity of a logical relationship (represented by formula $f$) between events. One possible approach could be to add an extension of $f$ to the premise. This reduces the inference problem to identifying whether the indicator event is identically true, identically false, or undefined within the allowed state space.

\subsection{Relation to Symbolic Logic}\label{subsec:relation-to-symbolic-logic}

\subsubsection{Logical Inference in the State Algebra}

A problem of logical inference is typically formulated as proving (or disproving) a logical sentence called a conclusion, given a set of logical sentences called a premise.\footnote{In propositional logic, we use the terms logical formula and logical sentence interchangeably.}

State Algebra offers a way to reframe logical inference using set operations on state vectors. If we assign a state vector to each sentence in the premise, the set of possible states satisfying all logical sentences of the premise can be determined by finding the intersection of these individual state vectors, which are treated as sets of states:
\[
    \bs^{(1)} \cap \bs^{(1)} \cap \ldots \cap \bs^{(m)}
\]
where $\bs^{(i)}$ is the state vector corresponding to the $i$-th formula in the premise, and $m$ is the total number of sentences in the premise.\footnote{The use of an index in parentheses here is to distinguish it from a different indexing notation used elsewhere.} Finding the total space of the allowed states is at the heart of many logical inference problems.

The correspondence between logical rules and state vectors allows us to change the language of logical inference from the symbolic notation of logical formulas to algebraic operations on state vectors. A large part of this paper is dedicated to algebraic operations on state vectors and state-algebraic algorithms for logical inference.

\subsubsection{Information State Vectors}\label{subsubsec:information_s_vectors}

A valid state vector corresponding to a logical formula is a set of states that satisfy the formula. An alternative and equivalent representation of the logical formula is a set of states that falsify the formula (prohibited states). State vectors interpreted as collections of prohibited states are called \textbf{information} state vectors.

We use hatted bold letters to represent information state vectors and square brackets for their content as follows:
\[
    \hat \bs = \begin{bmatrix}
                   1 & - & 0  \\
                   0 & 1 & -
    \end{bmatrix}
\]

The relationship between allowed and prohibited representations is straightforward: one is a complement to the other. If $\bs$ is an allowed state vector of a formula $f$, than the prohibited counterpart can be obtained as
\[
    \hat \bs = \bt \smallsetminus \bs
\]
where $\smallsetminus$ denotes a set difference. For instance, a formula $f \deff (E_1 \rightarrow E_2)$ can be represented as either of the two state vectors
\begin{align*}
    & \bs = \begin{Bmatrix}
                1 & 1 & - & \ldots & -\\
                0 & - & - & \ldots & -
    \end{Bmatrix}\\
    & \hat \bs = \begin{bmatrix}
                     1 & 0 & - & \ldots & -
    \end{bmatrix}
\end{align*}
A state is prohibited if it is prohibited by any formula in the premise. Thus, the state vector corresponding to the prohibited states of the premise is the union of the information state vectors representing individual sentences in the premise:
\[
    \hat \bs^{(1)} \cup \hat \bs^{(1)} \cup \ldots \cup \hat \bs^{(m)}
\]

\subsubsection{Mapping Between Symbolic Logic and State Vectors}

There is a one-to-one mapping between symbolic logical formulas and state vectors. Indeed, we already know how to construct a state vector that corresponds to any formula. Inverse mapping is easy to find if we notice that every row in a state vector is equivalent to a conjunction of literals and the entire state vector is a disjunction of its rows. Thus, a state vector can be equivalently represented as a symbolic formula in Disjunctive Normal Form (DNF). For instance
\begin{align*}
    & \bs =\begin{Bmatrix}
               1 & 0 & - & - & \ldots & -\\
               0 & - & 1 & - & \ldots & -
    \end{Bmatrix}  \quad \sim \quad
    (E_1 \wedge \lnot E_2)\vee (\lnot E_1 \wedge  E_3)
\end{align*}

Similarly, we notice that an information vector is a conjunction of
rows, whereas each row is a disjunction of negated literals, for example,
\begin{align*}
    & \hat \bs = \begin{bmatrix}
                     0 & - & 1 & - & \ldots & -\\
                     - & 1 & 0 & - & \ldots & -\\
    \end{bmatrix} \quad \sim \quad
    (E_1 \vee \lnot E_3) \wedge (\lnot E_2 \vee E_3)
\end{align*}
Thus, an information state vector can be equivalently represented as a symbolic formula in Conjunctive Normal Form (CNF).

The transformation between CNF and DNF in symbolic logic is equivalent to set subtraction $\bs = \bt \smallsetminus \hat \bs$ in the state vector representation.\footnote{This suggests a relatively simple way of constructing a state vector corresponding to a symbolic logical formula in CNF form by first computing $\hat\bs$ and then subtracting it from $\bt$. Of course, an allowed state vector corresponding to a CNF logical formula can as well be constructed directly like for any other logical formula.}

Using the correspondence between state vectors and logical formulas, we find that the atomic reduction of a state vector corresponds to the Shannon expansion in symbolic logic. For example, the following atomic reduction:
\[
    \begin{Bmatrix}
        0 & 1 & 0 & - & \ldots & -\\
        1 & 1 & 0 & - & \ldots & -
    \end{Bmatrix} =
    \begin{Bmatrix}
        - & 1 & 0 & - & \ldots & -
    \end{Bmatrix}
\]
corresponds to the following identity
\[
    (\lnot E_1 \wedge E_2 \wedge \lnot E_3) \vee (E_1 \wedge E_2 \wedge \lnot E_3)
    \quad \iff \quad (E_2 \wedge \lnot E_3)
\]

The atomic reduction in the information vector corresponds to the resolution principle (a dual form of Shannon expansion) in symbolic logic. For instance, atomic reduction
\[
    \begin{bmatrix}
        0 & 0 & 1 & - & \ldots & -\\
        0 & 1 & 1 & - & \ldots & -
    \end{bmatrix} =
    \begin{bmatrix}
        0 & - & 1 & - & \ldots & -
    \end{bmatrix}
\]
corresponds to the following identity
\[
    (E_1 \vee E_2 \vee \lnot E_3)\wedge (E_1 \vee \lnot E_2 \vee \lnot E_3)
    \quad \iff \quad (E_1 \vee \lnot E_3)
\]

\subsection{Logical Inference as a Sequence}\label{subsec:logical-inference-as-a-series}

In this section, we throw a quick glance beyond the scope of the finite state space. It is obvious that a reduced representation allows to formulate problems for an infinite system of events, as long as the number of formulas and their support remain finite.\footnote{Indeed, only \bound columns of logical formulas have nontrivial contribution to the reduced state vectors representation, all other columns will remain trivial, and can be ignored.} Here, we take a step towards considering arbitrarily large state spaces and sets of logical rules.

Let us consider a system of $N$ events and a set of $m$ logical formulas $\{f_i\}$ defined on the state space $S_\mathcal{E}$. If $N$ is finite, then there are $2^N$ possible states and $2^{2^N}$ distinct state vectors, and hence, distinct logical formulas. We refer to the set of logical formulas (finite or infinite) as the \emph{knowledge base}:
\[
    \text{knowledge base}\quad =\quad \{f_i\}_{i=1}^m
\]
By definition, logical formulas can have only finite support. However, if we allow $N$ to be infinite, we can also consider an infinite knowledge base. In this case, logical inference cannot generally be performed in a finite number of steps, and an iterative process must be considered instead. This process may or may not converge. Practically, early stopping needs to be used by inference algorithms; for instance, if the goal of the inference is achieved or a limit on run time is reached.

Viewing inference as an iterative process is particularly relevant in higher-order logic contexts. In such systems, the Herbrand Expansion -- the set of all ground atoms and formulas -- can be infinite or even uncountable. This renders even the theoretical consideration of the entire expansion impossible, positioning the iterative method as a practical alternative.

A primary task underlying almost any logical inference problem is to find an allowed subset of the state space, that is, a subset consisting of states that satisfy all logical formulas of the premise. Below, we define a formal stepwise process for finding the total allowed space, which can be applied to a finite or infinite knowledge base.

\subsubsection{Valid Set}\label{subsubsec:valid-set}

First, we consider valid state vectors, that is, those that represent allowed states. For every formula $f$ we find a corresponding valid state vector:
\begin{align}
    f \sim \bs\,,\qquad \bs = \{s: f(s) = 1\}
\end{align}

The knowledge base is represented as a set of $m$ state vectors
\[
    \text{knowledge base}\quad = \quad \{\bs^{(i)}\}_{i=1}^m
\]
A \textbf{complete valid set} (or simply, a valid set), $\bV$, is a state vector comprising all states that satisfy every formula within the knowledge base
\[
    \bV = \{s:\quad \forall i\,,\; f_i(s) = 1\}
\]

A state vector which comprises states satisfying a pair of formulas is obtained as the intersection of individual state vectors:
\[
    f_1\sim\bs^{(1)}\,,\ f_2\sim\bs^{(2)} \quad \Rightarrow \quad (f_1\wedge f_2) \sim (\bs^{(1)}\cap \bs^{(2)})
\]
Thus, the complete valid set is the intersection of all valid state vectors in the knowledge base.

We introduce a sequence of valid sets $\{\bV^{(i)}\}_{i=0}^m$, such that the $0$-th element is a trivial vector, and $i$-th element is obtained as an intersection of $(i-1)$-th element with the $i$-th state vector:
\begin{align}
    &\bV^{(0)} = \bt  \nonumber \\
    &\bV^{(1)} = \bV^{(0)} \cap \bs^{(1)} \nonumber \\
    & \cdots \nonumber \\
    &\bV^{(i)} = \bV^{(i-1)} \cap \bs^{(i)} \\
    & \cdots \nonumber
\end{align}
The $i$-th element in this sequence represents a set of states that satisfy the first $i$ formulas in the knowledge base. The resulting vector comprises states satisfying all formulas, and hence represents the complete valid set
\begin{align}
    \bV = \bV^{(m)} = \bigcap\limits_1^m \,\bs^{(i)}
\end{align}
At every step the valid set $\bV^{(i)}$ is a superset of the complete valid set
\begin{align}
    \bV \subseteq \bV^{(i)} \,,\qquad \text{for}\quad i<m
\end{align}
If the knowledge base is infinite, the complete valid set is obtained as a limit, where $m\to\infty$:
\begin{align}
    \bV = \bV^{(\infty)}
\end{align}

\subsubsection{Information Set}\label{subsubsec:information-set}

As mentioned previously, information state vectors pose an equivalent representation of State Algebra. An information vector $\hat \bs$ corresponding to formula $f$ comprises all states falsifying the formula.
\begin{align}
    f \sim \hat\bs\,,\qquad \hat\bs = \{s: f(s) = 0\}
\end{align}
The vectors $\bs$ and $\hat \bs$ are complementary:
\begin{align}
    \bs \cup \hat \bs = \bt\,,\quad \bs \cap \hat\bs = \varnothing
\end{align}

The knowledge base can be represented in terms of the information state vectors:
\[
    \text{knowledge base}\quad = \quad \{\hat \bs^{(i)}\}_{i=1}^m
\]
A \textbf{complete information set} (or simply, an information set), $\hat\bI$, is a state vector which comprises all states falsifying at least one formula in the knowledge base:
\[
    \hat\bI = \{ s: \quad \exists i\,,\;f_i(s) = 0\}
\]

The information vector of a pair of formulas is obtained as the union of the individual information vectors as follows:
\[
    f_1\sim \hat \bs^{(1)}\,,\ f_2\sim\hat\bs^{(2)} \quad \Rightarrow
    \quad (f_1\wedge f_2) \sim (\hat\bs^{(1)}\cup \hat\bs^{(2)})
\]
Thus, the complete information vector is obtained as the union of all information vectors in the knowledge base.

We introduce a sequence of information sets $\{\hat \bI^{(i)}\}_{i=0}^m$ as follows: the $0$-th element is an empty set, and the $i$-th element is obtained as a union of $(i-1)$-th element with the $i$-th information vector:
\begin{align}
    & \hat\bI^{(0)} = \bzero  \nonumber \\
    & \hat\bI^{(1)} = \hat\bI^{(0)} \cup \hat \bs^{(1)} \nonumber \\
    & \cdots \nonumber \\
    & \hat\bI^{(i)} = \hat\bI^{(i-1)} \cup \hat \bs^{(i)} \\
    & \cdots \nonumber
\end{align}
The $i$-th element in this sequence contains all the states that falsify at least one of the first $i$ formulas.

At the end of the inference process, vector $\hat\bI^{(m)}$ equals the complete information set
\begin{align}
    \hat\bI = \hat\bI^{(m)} = \bigcup\limits_1^m\, \hat\bs^{(i)}
\end{align}
At each intermediate step, the information set $\hat\bI^{(i)}$ is a subset of the complete information set
\begin{align}
    \hat \bI^{(i)} \subseteq \hat\bI\,,\qquad\text{for}\quad i<m
\end{align}

For an infinite knowledge base, the complete information vector is a limit, where $m\to\infty$:
\begin{align}
    \hat\bI = \hat\bI^{(\infty)}
\end{align}

The name ``information'' is inspired by the fact that the information vector of a set of formulas is a union of individual information vectors. Metaphorically, the information communicated by a collection of formulas is the sum of the information chunks conveyed by each of the formulas. The trivial state vector $\bt$ conveys no information, since the corresponding information vector is empty: $\hat\bt = \bzero$.

Note that, according to the set-theoretical duality principle,
\begin{align}
    & \forall i \leq m: \nonumber \\
    & \hat \bs^{(i)} = \bt \smallsetminus \bs^{(i)} \\
    & \hat\bI^{(i)} = \bt \smallsetminus \bV^{(i)}\\
    \label{eq:duality_valid_information}
    & \hat\bI = \bt \smallsetminus \bV
\end{align}
These relations are valid for an arbitrary $m$, whether it is finite or infinite.

\section{Algebra of State Vectors}\label{sec:algebra-of-state-vectors}

Truth tables serve as a valuable educational tool in classical mathematical logic, elucidating concepts such as logical entailment and inference. However, their practical application in numerical problems of realistic scale is limited by their exponential increase in size, relative to the number of events.

State Algebra employs state vectors that act as truth tables for logical formulas. The reduced representation of these state vectors significantly enhances their practical applicability in real-world scenarios. This reduction requires encoding only a minimal amount of information, leading to manageable state vector sizes, simplified set operations, and ultimately, efficient numerical algorithms for logical inference.

When interpreted as sets of states, operations on state vectors mirror set operations, specifically, union, intersection, and subtraction. We examine these operations in detail and demonstrate how compact representations enable efficient algorithms within the State Algebra.

\subsection{Basic Operations}\label{subsec:basic-operations}

When interpreted as sets of states, state vectors allow for standard set operations such as union, intersection, and subtraction:
\begin{align*}
    \bs \cup \br\,, \qquad \bs \cap \br\,, \qquad
    \bs \smallsetminus \br
\end{align*}

Let us consider the following example. For two following state vectors
\begin{align*}
    & \br =
    \begin{Bmatrix}
        0 & 1 & -\\
        1 & 0 & -
    \end{Bmatrix}, \\
    & \bq =
    \begin{Bmatrix}
        1 & 1 & - \\
        0 & 1 & -
    \end{Bmatrix},
\end{align*}
the three set operations yield
\begin{align*}
    & \br \cup \bq =
    \begin{Bmatrix}
        0 & 1 & - \\
        1 & 0 & - \\
        1 & 1 & -
    \end{Bmatrix}, \\
    & \br \cap \bq =
    \begin{Bmatrix}
        0 & 1 & -
    \end{Bmatrix}, \\
    & \br \smallsetminus \bq =
    \begin{Bmatrix}
        1 & 0 & -
    \end{Bmatrix}.
\end{align*}

Note that the 3rd column is trivial in both vectors $\bq$ and $\br$, and it can be ignored when calculating the set operations. In the resulting state vectors, the 3rd column is also trivial. The same applies to any number of columns that are trivial in both operands.

To make the notation of state-algebraic operations more convenient, we introduce a few definitions here. In the next section (Sec.~\ref{subsec:coordinate-representation}), we consider an interpretation that takes into account possible state duplicates (coordinate representation). In Section~\ref{subsec:index_notation}, we introduce the index notation, which allows us to express state vectors and algebraic operations concisely.

\subsection{Coordinate Representation}\label{subsec:coordinate-representation}

In some algorithms, it is necessary to track the number of states in a state vector, which also accounts for the possible duplicates. For this purpose, we introduce a coordinate representation. It generalises the set representation by viewing the state space as a formal algebraic structure.

\subsubsection{Definition of the Coordinate Representation}

Consider a system of $N$ events $\mathcal{E} = \{E_1, \ldots,E_N\}$ and a corresponding state space $S_\mathcal{E}$ consisting of $M=2^N$ distinct states $\{s_i\}_{i=1}^M$. Let us introduce an $M$-dimensional space over a ring of integer numbers, such that the objects $\{s_i\}$ constitute a basis. The objects in this space are arbitrary linear combinations
\[
    \sum_k a^k\,s_k\,,\qquad a^k \in \mathbb{Z}
\]
with integer coefficients $a^k$. An arbitrary state vector can be represented by the coordinates $x^k,\; k=1,\ldots,M$:
\[
    \bs = (x^1, x^2, \ldots ,x^M) = \sum_{k=1}^M x^k\,s_k\,,\qquad x^k \in \mathbb{Z}
\]

We refer to this space as the \textbf{coordinate representation} of the State Algebra. A coordinate $x^k$ is effectively a counter of how many times the state $s_k$ appears in $\bs$. For this reason, we sometimes refer to the coordinates as \textbf{multiplicity factors}. States that are not present in the matrix notation have a coordinate of 0 in the coordinate representation. Importantly, the coordinates are arbitrary integers. In particular, they can be negative or greater than 1 in magnitude.

Consider an example. For the set of two events $\{E_1, E_2\}$ the full state space consists of four states. We enumerate them as
\[
    \begin{array}{ccc}
        1: & 0 & 0\\
        2: & 0 & 1\\
        3: & 1 & 0\\
        4: & 1 & 1
    \end{array}
\]
Examples of the coordinate representation of the state vectors are as follows:
\[
    \begin{Bmatrix}
        0 & 0 \\
        1 & 1
    \end{Bmatrix} = \begin{pmatrix}
                        1 & 0 & 0 & 1
    \end{pmatrix}, \qquad
    \begin{Bmatrix}
        0 & 1 \\
        1 & 0 \\
        - & 1
    \end{Bmatrix} = \begin{pmatrix}
                        0 & 2 & 1 & 1
    \end{pmatrix}
\]
Note that in the second state vector, the state $\{0\ 1\}$ is duplicated; hence, the corresponding coordinate of the state vector equals~2.

An empty vector $\bzero$ has all its coordinates equal to 0. We use scalar 0 to designate an empty state vector, where it is unambiguous.
\begin{align*}
    & \bzero = (0, 0, \ldots,0)
\end{align*}
We will use interchangeably notation 1 or $\bt$ to designate the trivial state vector with all coordinates equal to 1:
\begin{align*}
    & 1 = (1, 1, \ldots,1)
\end{align*}

\subsubsection{Coordinate-wise Operations}

In the coordinate representation, we can introduce coordinate-wise operations: multiplication, addition, and subtraction.
Suppose we have two state vectors
\begin{align*}
    & \bs = (x^1, x^2,\ldots,x^M)\,,\\
    & \bq= (y^1, y^2,\ldots,y^M)\,.
\end{align*}
We define multiplication
\[
    \bs \, \bq= (x^1\, y^1, x^2\,y^2,\ldots, x^M\,y^M)\,,
\]
addition
\[
    \bs + \bq = (x^1 + y^1, x^2 + y^2,\ldots, x^M + y^M)\,,
\]
and subtraction
\[
    \bs - \bq = (x^1 - y^1, x^2 - y^2,\ldots, x^M - y^M)\,.
\]
These operations are analogous to the set operations of intersection, union, and set subtraction, but there are some important differences (see Sec.~\ref{subsubsec:binary_projection}).

For completeness, we can also introduce the multiplication by an integer scalar:
\begin{align}
    & a\,\bs = (a\,x^1, a\,x^2,\ldots, a\,x^M)\,,\qquad a\in \mathbb{Z}\,,
\end{align}
Obviously, addition and multiplication are commutative and distributive, for example,
\begin{align*}
    & \bs \,\bq = \bq\,\bs \\
    & \bs + \bq = \bq + \bs \\
    & \bs \,(\bq + \br) = \bs\,\bq + \bs\,\br
\end{align*}
and multiplication is bilinear. With these definitions, the space of the state vectors becomes an algebra over the ring of integers. We designate this space as
\[
    \mathbb{S}_\mathcal{E}
\]
The objects $\{\be_{(j)}\}$ defined as
\begin{align*}
    & \be_{(j)} = (\delta^i_{j})_{i=1}^M\,,\qquad \text{where} \quad
    \delta^i_{j} = \begin{cases}
                       1\,,\quad & i = j \\
                       0\,,      & \text{otherwise}
    \end{cases}
\end{align*}
constitute the basis in $\mathbb{S}_\mathcal{E}$.

We notice the following obvious relations
\begin{align}
    & 1 \cdot\bs = \bs \\
    & \bzero \cdot \bs = \bzero \\
    & 1 \cdot \bzero = \bzero  \\
    & \bs + \bzero = \bs \\
    & \bs - \bzero = \bs \\
    & \bs - \bs = \bzero
\end{align}
In the coordinate representation, two state vectors are considered \textbf{orthogonal} if their product is zero, a definition analogous to that used for orthogonality in set representation.
\begin{align*}
    \bs \perp \bq \qquad \iff \qquad \bs\,\bq = \bzero
\end{align*}
Note that, from an algebraic perspective, orthogonal state vectors are zero divisors.

State vectors are called \textbf{identical} if they have the same coordinate representation. For instance, the following three vectors are identical:
\[
    \begin{Bmatrix}
        1 & 1 \\
        0 & 1 \\
        0 & 0
    \end{Bmatrix} =
    \begin{Bmatrix}
        - & 1 \\
        0 & 0
    \end{Bmatrix} =
    \begin{Bmatrix}
        1 & 1 \\
        0 & -
    \end{Bmatrix}
\]

\subsubsection{Solving Equations}

In set representation, it is impossible to solve the equation
\[
    \bs \cup \bq = \br
\]
with respect to $\bs$ because the union of sets loses some information about operands. In the coordinate representation, the linear equation can be solved algebraically as follows:
\[
    \bs + \bq = \br \qquad \iff \qquad \bs = \br - \bq
\]
However, because multiplication has no inverse, it is generally not possible to solve the following equation with respect to $\bs$:
\[
    \bs\,\bq = \br
\]

Another important point which needs to be considered is that the existence of zero divisors prevents the cancellation property. For instance, in the equation
\[
    \bs\,\bq = \bs\,\br
\]
we cannot cancel $\bs$ (assuming $\bs\not = 0$) and conclude that $\bq=\br$. Instead, we can rewrite the equation as
\[
    \bs\,(\bq-\br) = 0
\]
which means $\bs$ and $\bq-\br$ must be orthogonal. This equation can have numerous solutions, and $\bq=\br$ is only one of them.

\subsection{Set Representation Versus Coordinate Representation}\label{subsec:set-vs-coordinate-representation}

Having established the set and coordinate representations of State Algebra, we now highlight the key differences and introduce some notational conventions.

\subsubsection{Set Representation as Equivalence Class}

Recall that the coordinate representation allows for integer-valued multiplicity factors, including negative ones. Conversely, a set representation treats state vectors as sets of states, ignoring duplicates and not attributing meaning to negative multiplicities.

Therefore, in a set representation, a state vector can be interpreted as an \emph{equivalence class}, where the equivalence relation is defined as follows: any state with a multiplicity factor greater than one is considered equivalent to a state with a multiplicity of one. This definition can be further extended by postulating the equivalence between a negative multiplicity and a multiplicity of zero.

Formally, two state vectors $\bs$ and $\bq$ are considered equivalent if, for all $i$:
\begin{align*}
    & \bs = (x^1, x^2,\ldots, x^M) \\
    & \bq = (y^1, y^2,\ldots, y^M) \\
    & \bs \simeq \bq\quad\iff \quad \forall i\,,\quad
    (x^i \leq 0  \Leftrightarrow y^i \leq 0) \wedge
    (x^i \geq 1  \Leftrightarrow y^i \geq 1)
\end{align*}
where we used symbol
\[
    \simeq
\]
for equality up to equivalence transformation.

Although, strictly speaking, state vectors in set and coordinate representations should not appear in the same algebraic expression, we will occasionally use such expressions when their interpretation is unambiguous. For example, if $\bs$ is in the set representation and $\bq$ is in the coordinate representation, the vector $\bq$ defines an equivalence class as $Q = \{\bx: \bx \simeq \bq\}$. In such cases, we will interpret the equation
\[
    \bs = \bq
\]
as a shortcut for the equation
\[
    \bs = Q
\]

\subsubsection{Binary Projection}\label{subsubsec:binary_projection}

Among all the equivalent forms of a state vector in the coordinate representation, one is of particular interest: when all the coordinates are exclusively either zero or one. We call this state vector \textbf{binary}. We also say that this vector has binary coordinates. Next, we introduce a \textbf{binary projection} operator $[\ ]$ which selects out of all elements of the equivalence class, the one with binary coordinates. If vector $\bs$ has coordinates $\bs = (x^1,x^2,\ldots,x^M)$, then vector $[\bs]$ has coordinates
\begin{align}
[\bs]
    = (y^i), \quad y^i =
    \begin{cases}
        1\,,\quad & \text{if  } x^i \geq 1 \\
        0\,,      & \text{if  } x^i \leq 0
    \end{cases}
\end{align}
Obviously, the operator $[\ ]$ is a projector: $[[\bs]] = [\bs]$. The state vector $\bs$ is binary if $[\bs] = \bs$. We explicitly mention whether a state vector is supposed to be binary.

The product of binary state vectors is binary, and the square of a binary state vector equals itself:
\begin{align}
    & (\text{binary }\bs,\bq) \nonumber \\
    & [\bs\,\bq] = \bs \,\bq \\
    & [\bs\,\bs] = \bs\,\bs = \bs
\end{align}

For binary state vectors, the following relationships are useful:
\begin{align}
    & (\text{binary }\bs,\bq) \nonumber \\
    & [\bs+\bq]  = \bs + \bq - \bs\, \bq \\
    & [\bs-\bq] = \bs - \bs\,\bq
\end{align}
It is easy to see why these identities hold if we notice that the product $\bs\,\bq$ contains all the states present in the intersection of the state vectors $\bs$ and $\bq$ considered as sets of states.

There is a natural one-to-one mapping between the state vectors in the set representation and the binary state vectors in the coordinate representation.

Generally, set operations, such as union $\cup$ or intersection $\cap$, are not defined in coordinate representation. However, one-to-one mapping allows for their interpretation when the operands (and the result) are binary state vectors. Binary projection helps establish a relationship between set operations $\cup$, $\cap$, $\smallsetminus$ and coordinate-wise operations $+$, $-$ and $\times$. For binary vectors $\bs$ and $\bq$, we define the following set operations:
\begin{align}
    & (\text{binary }\bs,\bq) \nonumber \\
    \label{eq:cross_binary}
    & \bs \cap \bq = \bs\,\bq \\
    \label{eq:union_binary}
    & \bs \cup \bq = [\bs + \bq] = \bs + \bq - \bs\, \bq \\
    \label{eq:minus_binary}
    & \bs \smallsetminus \bq = [\bs - \bq] =  \bs - \bs\,\bq
\end{align}
These operations are equivalent to set operations in the set representation. For some special cases, we get
\begin{align*}
    & (\text{binary }\bs) \\
    & 1 \cup \bs = [1 + \bs] = 1 + \bs - 1\cdot\bs  = 1 \\
    & 0 \cup \bs = [0 + \bs] = 0 + \bs - 0\cdot \bs = \bs \\
    & \bs \smallsetminus 1 = [\bs - 1] = \bs - 1\cdot\bs = 0 \\
    & \bs \smallsetminus 0 = [\bs - 0] = \bs - 0\cdot\bs = \bs \\
    & 1 \smallsetminus \bs = [1 - \bs] = 1 - 1\cdot\bs = 1 - \bs
\end{align*}

To illustrate how the mapping between set-theoretical and algebraic operations works, we consider the following example: suppose the knowledge base consists of two formulas that have the corresponding binary state vectors $\bs$ and $\bq$. The valid set is obtained as an intersection
\begin{align*}
    & \bV = \bs \cap \bq = \bs\,\bq
\end{align*}
The corresponding information vectors are
\[
    \hat \bs = 1 - \bs\,,\quad \hat \bq = 1 - \bq
\]
Evidently, these are also binary. The information set is obtained as
\begin{align*}
    & (\text{binary }\bs, \hat\bs, \bq, \hat\bq)\\
    \hat \bI & =  \hat \bs \cup \hat \bq =  \hat \bs + \hat\bq - \hat\bs\,\hat\bq  \\
    & = 1 - \bs + 1 - \bq - (1 - \bs)(1 - \bq)  = 1 - \bs\,\bq \\
    & = 1 - \bV  = 1 \smallsetminus \bV
\end{align*}
where the last equality is the general relation between the valid and information sets (Eq.~\ref{eq:duality_valid_information}).

\subsubsection{Generalisation of Some Definitions}

Binary projection allows us to extend some definitions introduced for state vectors in the set representation to the coordinate representation.

\begin{itemize}
    \item We say that an event $E$ is identically true (false) in the state vector $\bs$ if it is identically true (false) in the binary projection $[\bs]$.
    \item An event is \bound (\free) in the state vector $\bs$ if it is \bound (\free) in the binary projection~$[\bs]$.
    \item Support of a state vector is defined as support of its binary projection:
    \[
        \supp(\bs) = \supp([\bs])
    \]
    \item Two state vectors are independent if their binary projections are independent:
    \[
        \bs \wr \bq \qquad \iff \qquad [\bs] \wr [\bq]
    \]
\end{itemize}

\subsection{Index Notation}\label{subsec:index_notation}

Here, we introduce a notation which simplifies algebraic expressions and allows the algorithms to be formulated in a concise manner.

\subsubsection{Subvector}

In matrix notation, the index is used to denote the subvectors of the state vector. An upper index indicates a subvector containing only states in which the value of the corresponding event is one, whereas a lower index indicates a subvector in which the value of the corresponding event is zero. For example:
\[
    \bs = \begin{Bmatrix}
              1 & 1 & - \\
              0 & - & 0 \\
              - & 0 & 1
    \end{Bmatrix}\,;\qquad
    \bs^1 = \begin{Bmatrix}
                1 & 1 & - \\
                1 & 0 & 1
    \end{Bmatrix}\,;\qquad
    \bs_3 = \begin{Bmatrix}
                1 & 1 & 0 \\
                0 & - & 0
    \end{Bmatrix}
\]
In the coordinate representation, an upper index $i$ nullifies the coordinates of all states in which $e_i=0$ (where $e_i$ is the value of event $E_i$). Conversely, a lower index $i$ sets to zero the coordinates of all states in which $e_i=1$. If the state vector $\bs$ has coordinates $\bs = (x^1, x^2, \ldots,x^M)$, then the coordinates of the vector $\bs^i$ are defined as
\begin{align*}
    \bs^i = (y^k)_{k=1}^M\,,\quad y^k =
    \begin{cases}
        x^k\,, & \text{if  } e_{ki} = 1 \\
        0\,,        & \text{if  } e_{ki} = 0
    \end{cases}
\end{align*}
Here, $e_{ki}$ represents the value of the $i$-th event in the $k$-th state. Similarly, the coordinates of the state vector with lower index are defined as
\begin{align*}
    \bs_i = (y^k)_{k=1}^M\,,\quad y^k =
    \begin{cases}
        0\,, & \text{if  } e_{ki} = 1 \\
        x^k\,,        & \text{if  } e_{ki} = 0
    \end{cases}
\end{align*}
A trivial state vector with an index has a single entry of 0 or 1, and holes otherwise; for instance,
\begin{align*}
    &\bt_1 = \begin{Bmatrix}
                 0 & - & - &\ldots & -
    \end{Bmatrix}\,,\qquad
    \bt^2 = \begin{Bmatrix}
                - & 1 & - &\ldots & -
    \end{Bmatrix}
\end{align*}

Note that the state vector $\bt^n$ has coordinates 1 for all states where $E_n$ is true and 0 if $E_n$ is false. Therefore, a subvector $\bs^n$ (or $\bs_n$) can be obtained by multiplying with $\bt^n$ (or $\bt_n$):
\begin{align}
    \label{eq:def_s^n}
    \bs^n = \bs\,\bt^n\,, \qquad \bs_n = \bs\,\bt_n
\end{align}

The operation for obtaining a subvector distributes
\begin{align}
(\bs + \bq)
    _n = (\bs + \bq)\,\bt_n = \bs\,\bt_n + \bq\,\bt_n = \bs_n + \bq_n
\end{align}
By observing that $\bt_n\,\bt_n = \bt_n$ and $\bt^n\,\bt^n = \bt^n$, we obtain the following useful identities:
\begin{align}
    & (\bs\,\bq)_n = \bs_n \,\bq_n = \bs\,\bq_n = \bs_n\,\bq
\end{align}

\subsubsection{Atomic Reduction}

Notice that an atomic reduction by index $n$ of the following two rows
\[
    \begin{Bmatrix}
        - & \ldots & - & 0 & - & \ldots & - \\
        - & \ldots & - & 1 & - & \ldots & -
    \end{Bmatrix} =
    \begin{Bmatrix}
        - & - & \ldots
    \end{Bmatrix}
\]
can be expressed in terms of trivial vectors as
\begin{align}
    \bt_n + \bt^n = \bt
\end{align}
We can use this identity to show that, for an arbitrary vector $\bs$,
\begin{align}
    \bs = \bs\,\bt = \bs\,(\bt_n + \bt^n) = \bs_n + \bs^n
\end{align}
A similar identity holds for the product of the state vectors:
\begin{align}
    & \bs\,\bq = \bs \, \bq\,(\bt_n + \bt^n) = \bs_n \,\bq_n +  \bs^n \,\bq^n
\end{align}

\subsubsection{Multiple Indices}

Consider a space of $N$ events. Let $\aleph$ be the set of all finite subsets of the set of integers $\{1, \ldots,N\}$. Let $\alpha=\{i_1,\ldots,i_m\}$ and $\beta=\{j_1,\ldots,j_n\}$ be finite sets of indices: $\alpha, \beta \in\aleph$. Then we interpret
\[
    \bs^\alpha_\beta
\]
as a state vector $\bs$ to which the operation subvector is applied for all indices in the sets $\alpha$ and $\beta$. The order in which the operation subvector is applied is irrelevant, since
\[
    \bs^{ij} = \bs\,\bt^i\,\bt^j = \bs\,\bt^j\,\bt^i = \bs^{ji}\,,\qquad (\bs^i)_j = (\bs_j)^i = \bs^i_j
\]
Hence, sets $\alpha$ and $\beta$ need not be ordered.

Subvector is a projection: repeating subvector operation has no effect
\begin{align}
    \bs^{n\,n} = \bs\,\bt^n\,\bt^n = \bs\,\bt^n = \bs^n
\end{align}

State vector with coinciding upper and lower index is empty
\begin{align}
    & \bs^i_i = \bzero\,, \\
    & \bs^\alpha_\beta = \bzero \iff \alpha\cap\beta \not = \varnothing
\end{align}

In the following, we employ a simplified notation for the index sets. Specifically, $\alpha_1\,\alpha_2$ represents the union of sets of indices $\alpha_1\cup\alpha_2$. Likewise, $\alpha\,ij$ denotes the union $\alpha \cup \{ij\}$. For example:
\[
    \bs^{\alpha_1\,\alpha_2}_{\beta\,ij} =  \bs^{\alpha_1\cup\alpha_2}_{\beta\cup\{ij\}}
\]

\subsection{T-objects}\label{subsec:t-objects}

A trivial vector with indices, referred to as a \textbf{t-object}, can be represented as a single row in matrix notation, composed of 0s and 1s at the corresponding positions of the indices, and holes in other positions. For example:
\[
    \bt^{14}_{36} = \begin{Bmatrix}
                        1 & - & 0 & 1 & - & 0 & - & - & \ldots
    \end{Bmatrix}
\]
T-objects in the coordinate representation can be added ($+$), subtracted ($-$), and multiplied~($\times$). Note that, by design, t-objects are \emph{binary}, and hence, inherently support set operations such as $\cap$, $\cup$, $\smallsetminus$, $\subseteq$. For instance:
\begin{align*}
    & \bt_{(1)} \cup \bt_{(2)} = \bt_{(1)} + \bt_{(2)} - \bt_{(1)}\,\bt_{(2)} \\
    & \bt_{(1)} \subseteq \bt_{(2)} \qquad \iff \qquad \bt_{(1)} = \bt_{(1)}\,\bt_{(2)}
\end{align*}
Given that t-objects are binary, their coordinates are always nonnegative. This leads to a simple yet important conclusion: the sum of the nonzero t-objects can never be equal to zero.
\begin{align}
    \label{eq:sum-of-t-objects}
    \bt_{(1)} + \bt_{(2)} + \ldots = \bzero \quad \iff \quad \forall i,\ \bt_{(i)} = \bzero
\end{align}

\subsubsection{Multiplication}

Below, we suppose that $\alpha_1,\beta_1,\alpha_2, \beta_2 \in\aleph$.

It is easy to see how t-objects with indices can be multiplied.
\begin{align}
    \bt^{\alpha_1}_{\beta_1}\, \bt^{\alpha_2}_{\beta_2} = \bt^{\alpha_1\,\alpha_2}_{\beta_1\,\beta_2}
\end{align}
For example
\begin{align*}
    & \bt^{1}_{2} \,\bt^{5}_{6} = \bt^{15}_{26} \\
    & \bt^{1}_{2} \,\bt^{15}_{26} = \bt^{15}_{26}
\end{align*}

For an arbitrary state vector $\bs$ we can write
\begin{align}
    & \bs^{i_1, \ldots, i_m}_{j_1,\ldots,j_n} = \bs\,\bt^{i_1}\ldots \bt^{i_m}\,\bt_{j_1}\ldots\bt_{j_n} =
    \bs\,\bt^{i_1, \ldots, i_m}_{j_1,\ldots,j_n}\,,
\end{align}
consequently:
\begin{align}
    & (\text{binary } \bs) \nonumber \\
    & \bs^{\alpha_1}_{\beta_1}\, \bs^{\alpha_2}_{\beta_2} =
    \bs\,\bs\, \bt^{\alpha_1}_{\beta_1}\, \bt^{\alpha_2}_{\beta_2} =
    \bs\,\bt^{\alpha_1\,\alpha_2}_{\beta_1\,\beta_2} =
    \bs^{\alpha_1\,\alpha_2}_{\beta_1\,\beta_2}\,.
\end{align}

Atomic reduction with multiple indices becomes
\begin{align}
    & \bt^{\alpha \,i}_{\beta} + \bt^{\alpha}_{\beta\,i} =
    \bt^\alpha_\beta\,(\bt^i + \bt_i) = \bt^\alpha_\beta
\end{align}
Similarly, for an arbitrary vector
\begin{align}
    & \bs^{\alpha\,i}_{\beta} + \bs^{\alpha}_{\beta\,i}=
    \bs^\alpha_\beta\,(\bt^i + \bt_i) = \bs^\alpha_\beta
\end{align}

\subsubsection{Subset}\label{subsubsec:subset}

For arbitrary $\alpha_1,\beta_1,\alpha_2, \beta_2$, vector $\bt^{\alpha_1 \,\alpha_2}_{\beta_1\,\beta_2}$ is a subset of $\bt^{\alpha_1}_{\beta_1}$
\[
    \bt^{\alpha_1\,\alpha_2}_{\beta_1\,\beta_2} \subseteq \bt^{\alpha_1}_{\beta_1}
\]
This is a consequence of the fact that their intersection coincides with the left operand:
\begin{align*}
    & \bt^{\alpha_1\,\alpha_2}_{\beta_1\,\beta_2}\cap\bt^{\alpha_1}_{\beta_1} =
    \bt^{\alpha_1\,\alpha_2}_{\beta_1\,\beta_2}\,\bt^{\alpha_1}_{\beta_1} =
    \bt^{\alpha_1\,\alpha_2}_{\beta_1\,\beta_2}
\end{align*}

If $\alpha_1\cap\beta_1= \varnothing$ and at least one of
$(\alpha_2\smallsetminus \alpha_1)$ or $(\beta_2\smallsetminus \beta_1)$
is nonempty, then $\bt^{\alpha_1\,\alpha_2}_{\beta_1\,\beta_2}$ is a proper subset of $\bt^{\alpha_1}_{\beta_1}$.

This justifies the intuition regarding the following useful relations:
\begin{align}
    \label{eq:reducdent_states}
    & [\bt^{\alpha_1\,\alpha_2}_{\beta_1\,\beta_2} + \bt^{\alpha_1}_{\beta_1}] =
    \bt^{\alpha_1\,\alpha_2}_{\beta_1\,\beta_2} + \bt^{\alpha_1}_{\beta_1} -
    \bt^{\alpha_1\,\alpha_2}_{\beta_1\,\beta_2} \, \bt^{\alpha_1}_{\beta_1}
    =\bt^{\alpha_1}_{\beta_1} \\
    & [\bt^{\alpha_1\,\alpha_2}_{\beta_1\,\beta_2} - \bt^{\alpha_1}_{\beta_1}] =
    \bt^{\alpha_1\,\alpha_2}_{\beta_1\,\beta_2} -
    \bt^{\alpha_1\,\alpha_2}_{\beta_1\,\beta_2} \, \bt^{\alpha_1}_{\beta_1}=0
\end{align}

The same applies to an arbitrary binary state vector $\bs$:
\begin{align}
    & (\text{binary }\bs) \nonumber \\
    & [\bs^{\alpha_1\,\alpha_2}_{\beta_1\,\beta_2} + \bs^{\alpha_1}_{\beta_1}]
    =\bs^{\alpha_1}_{\beta_1} \\
    & [\bs^{\alpha_1\,\alpha_2}_{\beta_1\,\beta_2} - \bs^{\alpha_1}_{\beta_1}] = 0
\end{align}

\subsubsection{Orthogonality}

Two t-objects are \textbf{orthogonal} if their product vanishes. This occurs if they have overlapping upper and lower indices:
\[
    \bt^{\alpha_1}_{\beta_1}\,\bt^{\alpha_2}_{\beta_2} = \bt^{\alpha_1\,\alpha_2}_{\beta_1\,\beta_2} = 0
    \qquad \iff \qquad
    (\alpha_1\cup \alpha_2)\cap(\beta_1\cup\beta_2) \not = \varnothing
\]
For instance,
\[
    \bt^{3}_{2} \,\bt^{5}_{23} = \bt^{35}_{23}  = 0
\]
This expression vanishes owing to the coinciding upper and lower indices of 3.

\subsubsection{Orthogonalisation}\label{subsubsec:orthogonalisation}

Consider two t-objects
\[
    \bt_{(1)} = \bt^{\alpha_1}_{\beta_1}\,,\qquad \bt_{(2)} = \bt^{\alpha_2}_{\beta_2}
\]
Given that t-objects are binary, they can be viewed as a set of states. In this section, we will demonstrate how $\bt_{(1)}$ can be decomposed into two distinct sets of states: a collinear component~$\bT_{\parallel}$, comprising states also found in $\bt_{(2)}$, and an orthogonal component $\bT_{\perp}$, consisting of states not present in $\bt_{(2)}$.

Let us designate
\[
    \gamma = \alpha_2 \smallsetminus \alpha_1\,,\qquad \nu = \beta_2 \smallsetminus \beta_1
\]
For the sake of demonstration, let's assume that the sets $\gamma$ and $\nu$ each contain two indices
\[
    \gamma = \{ij\}\,,\quad \nu = \{lm\}
\]
Generalisation to any other combination of indices would be straightforward. It is easy to see that the vector $\bt_{(1)}$ can be represented as
\begin{align*}
    & \bt^{\alpha_1}_{\beta_1} =
    \bt^{\alpha_1\,ij}_{\beta_1\,lm} +
    \bt^{\alpha_1\,i}_{\beta_1\,lmj} +
    \bt^{\alpha_1}_{\beta_1\,lmi} +
    \bt^{\alpha_1\,m}_{\beta_1\,l} +
    \bt^{\alpha_1\,l}_{\beta_1}
\end{align*}
To demonstrate why this identity holds, we can recursively apply atomic reduction to the leftmost terms. For instance, the first two terms can be reduced by index $j$ to yield
\[
    \bt^{\alpha_1\,ij}_{\beta_1\,lm} + \bt^{\alpha_1\,i}_{\beta_1\,lmj} = \bt^{\alpha_1\,i}_{\beta_1\,lm}
\]
The result of the first reduction can be reduced with the next term to eliminate index $i$, and so on, until all indices $\{i, j, l, m\}$ are eliminated.

The first term in this series is the collinear component
\begin{align*}
    &\bT_{\parallel} = \bt_{(1)}\,\bt_{(2)} = \bt^{\alpha_1\,\alpha_2}_{\beta_1\,\beta_2} =
    \bt^{\alpha_1\,ij}_{\beta_1\,lm}
\end{align*}
Consequently, the rest of the series is the orthogonal component
\begin{align*}
    &\bT_{\perp} = \bt^{\alpha_1\,i}_{\beta_1\,lmj} +
    \bt^{\alpha_1}_{\beta_1\,lmi} +
    \bt^{\alpha_1\,m}_{\beta_1\,l} +
    \bt^{\alpha_1\,l}_{\beta_1}
\end{align*}
such that
\[
    \bt_{(1)} = \bT_{\parallel} + \bT_{\perp}
\]

We now prove some properties of the collinear and orthogonal components. First, we notice that both are binary and mutually orthogonal:
\begin{align*}
    & [\bT_{\parallel}] = \bT_{\parallel} \\
    & [\bT_{\perp}] = \bT_{\perp} \\
    & \bT_{\parallel}\, \bT_{\perp} = 0
\end{align*}
The binary nature of $\bT_{\perp}$ is a direct consequence of its construction as the sum of pairwise orthogonal t-objects. Next, we check that the set intersections are as expected:
\begin{align*}
    & \bt_{(1)}\,\bT_{\parallel} = \bT_{\parallel} \\
    & \bt_{(1)}\,\bT_{\perp} = \bT_{\perp} \\
    & \bt_{(2)}\,\bT_{\parallel} = \bT_{\parallel} \\
    & \bt_{(2)}\,\bT_{\perp} = 0
\end{align*}
These identities can be easily verified by direct substitutions.

\subsubsection{Subtracting t-objects}

The decomposition of a t-object into collinear and orthogonal components allows us to calculate a binary difference
between t-objects (i.e. a set difference of t-objects interpreted as sets of states). Indeed
\begin{align}
    & \bt_{(1)} \smallsetminus \bt_{(2)} = [\bt_{(1)} - \bt_{(2)}] =  \bt_{(1)} - \bt_{(1)}\,\bt_{(2)} =
    \bt_{(1)} - \bT_{\parallel} = \bT_{\perp}
\end{align}
or explicitly
\begin{align}
    \label{eq:subtract_t_t}
    & \bt_{(1)} \smallsetminus \bt_{(2)} =
    \bt^{\alpha_1\,i}_{\beta_1\,lmj} +
    \bt^{\alpha_1}_{\beta_1\,lmi} +
    \bt^{\alpha_1\,m}_{\beta_1\,l} +
    \bt^{\alpha_1\,l}_{\beta_1}
\end{align}

We want to highlight two limiting cases. First, if t-objects $\bt^{\alpha_1}_{\beta_1}$ and $\bt^{\alpha_2}_{\beta_2}$ are orthogonal, then the result of the subtraction equals the first operand:
\[
    \bt_{(1)}\,\bt_{(2)} = 0 \qquad \Rightarrow \qquad \bt_{(1)} \smallsetminus \bt_{(2)} = \bt_{(1)}
\]
On the other hand, if the first operand is a subset of the second, or equivalently, if $\gamma=\varnothing$ and $\nu=\varnothing$, then the result of the subtraction is empty:
\[
    \bt_{(1)}\,\bt_{(2)} = \bt_{(1)} \qquad \Rightarrow \qquad \bt_{(1)} \smallsetminus \bt_{(2)}=0\,.
\]

\subsection{Row Decomposition}\label{subsec:row-decomposition}

Every row of a state vector in matrix notation can be viewed as a t-object $\bt^{\alpha}_{\beta}$, where $\alpha$ and $\beta$ are nonoverlapping sets of indices. When interpreted as a set of states, the state vector is the union of its rows. For example:
\[
    \bs = \begin{Bmatrix}
              1 & 0 & - \\
              0 & - & 0
    \end{Bmatrix} = \bt^1_2 \cup \bt_{13}
\]

However, if we interpret the same matrix in the coordinate representation, it becomes a sum of t-objects representing its rows. We refer to this representation of a state vector as the \emph{row decomposition}. The row decomposition of the example above reads as follows:
\[
    \bs = \begin{Bmatrix}
              1 & 0 & - \\
              0 & - & 0
    \end{Bmatrix} = \bt^1_2 + \bt_{13}
\]
Row decomposition interprets the state vectors in a coordinate representation, thereby accounting for potential state duplicates. However, it is a truncated form of coordinate representation because it only allows nonnegative multiplicity factors. Formally, we define row decomposition as a space in which state vectors are represented as the sum of arbitrary t-objects:
\[
    \bs = \bt^{\alpha_1}_{\beta_1} + \bt^{\alpha_2}_{\beta_2} + \bt^{\alpha_3}_{\beta_3} + \ldots
\]
The sum is equivalent to an arbitrary linear combination with nonnegative coefficients. Coefficients larger than 1 can be represented by repeating the same term multiple times. Any state vector with nonnegative coordinates can be represented as a sum of t-objects. Consequently, row decomposition is a semi-algebra defined on a semi-ring of nonnegative integer numbers, allowing for addition ($+$) and multiplication ($\times$), but not for subtraction ($-$) or division~($\div$). The space of state vectors in the row decomposition representation is denoted by
\[
    \mathbb{S}^+_\mathcal{E}
\]
An \emph{equivalence transformation} in the row decomposition representation must preserve the nonnegativity of the multiplicity factors. We will be using symbol
\[
    \cong
\]
to denote equivalence which preserves coordinate nonnegativity.

With respect to subtraction, we can impose less stringent constraints. Namely, we can allow the subtraction of t-objects as long as the result of the subtraction can be represented as a row decomposition with nonnegative coefficients. For instance, in the equation $\bt_{(1)} + \bt_{(2)} = \bt_{(1)} + \bt_{(3)}$ we can subtract $\bt_{(1)}$ from both sides to obtain $\bt_{(2)} = \bt_{(3)}$.

The lack of negative multiplicity factors makes row decomposition easier to interpret from a set-theoretic perspective. In particular, we can say that a state is present or absent in a state vector based on whether its multiplicity factor is zero.

We notice the following simple but powerful statements about the row decomposition representation of State Algebra:
\begin{itemize}
    \item A state vector is empty if and only if it has zero t-objects. This statement does not hold in the coordinate representation which allows negative multiplicity factors. For example, as illustrated in Eq.~\eqref{eq:trivial_irreducible}, the following state vector is empty:
    \[
        \bt_{123} + \bt^{123} + \bt^2_1 + \bt^3_2 + \bt^1_3 - \bt = 0
    \]
    However, the sum of t-objects cannot be empty.

    Some algorithms require proving that a state vector is empty. The statement above allows us to have a simple criterion for an empty state vector: it has to have zero t-objects.
    \item If an event is identically true (or false) in a state vector in row decomposition representation, it remains identically true (or false) under equivalence transformation preserving nonnegativity of multiplicity factors.
\end{itemize}
These statements are useful in some inference algorithms because they justify why the results are invariant under equivalence transformations.

\subsubsection{Row Decomposition of Binary Vectors}
Although trivial, the following statement is important for practical implementations: a state vector in row decomposition is binary if and only if it is represented as a sum of mutually orthogonal t-objects. Indeed, let
\[
    \bs = \sum_i a_i\,\bt_{(i)}\,,\quad a_i > 0
\]
The vector $\bs$ is binary if $\bs\,\bs = \bs$. The square of $\bs$ is given by
\[
    \bs\,\bs = \sum_i a_i^2\,\bt_{(i)} + \sum_{i \not =j} a_i\,a_j\,\bt_{(i)}\,\bt_{(j)}
\]
Substituting this expression into $\bs\,\bs = \bs$ and subtracting the right-hand side, we obtain
\[
    \sum_i (a_i^2 - a_i)\,\bt_{(i)} + \sum_{i \not =j} a_i\,a_j\,\bt_{(i)}\,\bt_{(j)} = 0
\]
We are allowed to do subtraction because $a_i^2 - a_i \geq 0$ for natural numbers. Recall that the sum of t-objects cannot be zero. Hence, we conclude that
\begin{align*}
    & a_i = 1\,,\quad  \forall i \\
    & \bt_{(i)}\,\bt_{(j)} = 0\,,\quad \forall i\not = j
\end{align*}
Note that the operations of multiplication, reduction, and taking a subvector preserve the binary property. If the knowledge base is represented as a set of binary vectors, all state vectors obtained in the process of logical inference remain binary as long as only these three operations are used. If a state vector is nonbinary, it can be transformed into an equivalent binary form by means of orthogonalisation (see Sec.~\ref{subsubsec:orthogonalisation}).

\section{Algebra of t-objects}\label{sec:algebra-of-t-objects}

State Algebra was introduced as a formalisation of the set-theoretical view of a state space -- a space of possible attributions of truth values to a set of events.

In this section, we define State Algebra as a space of abstract t-objects and operations on them, without anchoring to the semantic interpretation of t-objects as subsets of a truth table. The t-object formalism provides a framework for describing mathematical logic using abstract algebraic terms.

\subsection{Definitions}\label{subsec:definitions}
Let us consider a set of natural numbers $\mathbb{N} = \{1, 2,\ldots,N\}$ with $N$ being finite or infinite. Let $\mathbb{\aleph}$ be a set of all possible \emph{finite} subsets of $\mathbb{N}$. For the elements of $\aleph$ we use typical set operations: union, intersection, and subtraction. Note that the complement of an element of $\aleph$ is not an element of $\aleph$ if $N$ is infinite.

For arbitrary sets $\alpha,\beta\in\aleph$, we consider objects with upper and lower indices
\[
    \bt^{\alpha}_\beta\,,\qquad \alpha,\beta\in\aleph
\]
which we refer to as \textbf{t-objects}.

We denote the \textbf{unit} t-object $\bt$ with empty sets $\alpha$ and $\beta$ as \emph{one}
\[
    \bt = 1
\]
We also introduce a special \textbf{null} object $\bzero$. By definition, if sets $\alpha$ and $\beta$ overlap, then the t-object is null:
\[
    \bt^{\alpha}_\beta = \bzero \quad \text{iff}\quad \alpha\cap \beta \not = \varnothing
\]
We define \textbf{state vectors} as finite linear combinations of t-objects with integer coefficients
\[
    \bs = \sum_{i = 1}^k a_i\,\bt^{\alpha_i}_{\beta_i}\,,\qquad
    \alpha_i,\beta_i \in\aleph\,,\qquad a_i\in\mathbb{Z}\,,\qquad k<\infty
\]
State vectors form a $\mathbb{Z}$-module $\mathbb{S}$ with all the standard properties of a module: the addition of state vectors is a commutative group, scalar multiplication is associative, and distributes over addition. The null object plays the role of an additive identity element. By definition:
\begin{align}
    & \forall \bs : \nonumber \\
    & \bs + (-\bs) = \bzero \\
    & \bzero + \bs = \bs + \bzero = \bs \\
    & 0 \cdot\bs = \bzero
\end{align}
We will use scalar 0 to designate the null object, where it is unambiguous.

Objects $\bt^\alpha_\beta$ constitute the generating set in $\mathbb{S}$. However, the operation decomposition introduced below makes many t-objects to be oblique. In Section~\ref{subsec:finite-dimensional-space}, we focus on constructing an orthogonal basis for a finite-dimensional space.

\subsection{Multiplication}\label{subsec:multiplication}

We define a multiplication operation on t-objects as
\begin{align}
    \bt^{\alpha_1}_{\beta_1}\,\bt^{\alpha_2}_{\beta_2} =
    \bt^{\alpha_1\,\alpha_2}_{\beta_1\,\beta2}
\end{align}
where we use the simplified notation $\alpha_1\,\alpha_2 = \alpha_1 \cup\alpha_2$. Multiplication is obviously commutative. The unit object $\bt=1$ with empty sets $\alpha$ and $\beta$ plays the role of a multiplicative identity. The multiplication of t-objects is not equipped with an inverse.

The object $\bzero$ has a property
\begin{align}
    \bzero \cdot\bt^\alpha_\beta = \bt^\alpha_\beta \cdot \bzero = \bzero \qquad \forall \alpha, \beta
\end{align}
The object $1$ has a property
\begin{align}
    1 \cdot \bt^\alpha_\beta = \bt^\alpha_\beta \cdot 1 = \bt^\alpha_\beta \qquad \forall \alpha, \beta
\end{align}
The multiplication of t-objects is both commutative and associative. We define multiplication as being distributive over addition.
\begin{align}
    & (\bt_{(1)} + \bt_{(2)})\,\bt_{(3)} = \bt_{(1)}\,\bt_{(3)} + \bt_{(2)}\,\bt_{(3)} \\
    & \bt_{(1)}\,(\bt_{(2)} + \bt_{(3)}) = \bt_{(1)}\,\bt_{(2)} + \bt_{(1)}\,\bt_{(3)}
\end{align}
The multiplication of t-objects induces the multiplication of the state vectors.
\begin{align*}
    & \bs = \sum_i a_i\,\bt^{\alpha_i}_{\beta_i} \\
    & \bq = \sum_j b_j\,\bt^{\mu_j}_{\nu_j} \\
    & \bs\,\bq = \sum_i\sum_j a_i\,b_j\,\bt^{\alpha_i\mu_j}_{\beta_i\nu_j}
\end{align*}
The null and unit objects have the same effect on the state vectors:
\begin{align*}
    & \forall \bs: \\
    &\bzero \cdot\bs = \bs\cdot \bzero = \bzero \\
    & 1\cdot \bs = \bs\cdot 1 = \bs
\end{align*}

The multiplication of state vectors is commutative and has no inverse. With these definitions, $\mathbb{S}$ becomes an algebra over a ring of integer numbers.

\subsection{Orthogonality and Linear Independence}

T-objects are called \textbf{orthogonal} if their product vanishes:
\begin{align}
    \bt_{(1)} \perp \bt_{(2)} \qquad \iff\qquad \bt_{(1)}\,\bt_{(2)} = 0
\end{align}

The orthogonality of the state vectors is defined similarly:
\begin{align}
    \bs \perp \bq \qquad \iff \qquad \bs\,\bq = 0
\end{align}
Orthogonal vectors are zero divisors in $\mathbb{S}$.

A state vector is called \textbf{binary} if its square equals itself:
\[
    \bs\,\bs = \bs
\]
Any t-object is binary
\[
    \bt^\alpha_\beta\,\bt^\alpha_\beta =  \bt^{\alpha\,\alpha}_{\beta\,\beta} = \bt^\alpha_\beta
\]
If two t-objects are orthogonal, their sum is also binary:
\[
    (\bt_{(1)} + \bt_{(2)})(\bt_{(1)} + \bt_{(2)}) = \bt_{(1)} + 2\,\bt_{(1)}\,\bt_{(2)} + \bt_{(2)} =
    \bt_{(1)} + \bt_{(2)}
\]

A set of mutually orthogonal t-objects is linearly independent. Indeed, suppose the following linear combination vanishes
\[
    a_1\,\bt_{(1)} + a_2\,\bt_{(2)} + \ldots + a_n\,\bt_{(n)} = 0
\]
Multiplying both sides by $\bt_{(i)}$ and noticing that $\bt_{(i)}\,\bt_{(i)} = \bt_{(i)}$ and $\bt_{(i)}\,\bt_{(j)} = 0$ for $i\not = j$, we conclude that
\[
    \forall i,\quad  a_i = 0
\]

\subsection{Decomposition and Reduction}\label{subsec:decomposition-and-reduction}

We introduce the \textbf{decomposition by index} operation as follows: for arbitrary $\alpha,\beta\in\aleph$ and an arbitrary index $i \in\mathbb{N}$, the decomposition by index $i$ is an identity defined as
\begin{align}
    \bt^\alpha_\beta = \bt^{\alpha\,i}_\beta + \bt^\alpha_{\beta\,i}\,,\qquad
    \forall \alpha, \beta :\ \alpha\cap\beta=\varnothing\,,\qquad \forall i:\ i\not\in\alpha, \beta
\end{align}
where we used the simplified notation $\alpha\,i$ for $\alpha\cup\{i\}$. The same identity read from right to left is called the \textbf{atomic reduction by index}. Note that decomposition by index $i$ splits a t-object into a pair of new t-objects only if $i$ is not present in either $\alpha$ or $\beta$. Although the decomposition identity is formally valid even if $i$ is an element of either $\alpha$ or $\beta$, we always assume $i\not\in\alpha, \beta$.

We call the \textbf{descendants} of a t-object all nonempty t-objects that can be obtained by the recursive application of decomposition. For example, $\bt^{\alpha\,i}_\beta$ and $\bt^\alpha_{\beta\,i}$ are the descendants of $\bt^\alpha_\beta$. After applying the decomposition twice by indices $i$ and $j$, we identify another four descendants:
\[
    \bt^\alpha_\beta = \bt^{\alpha\,i\,j}_{\beta} +
    \bt^{\alpha\,i}_{\beta\,j} +
    \bt^{\alpha\,j}_{\beta\,i} +
    \bt^{\alpha}_{\beta\,i\,j}
\]

For the following discussion, it is useful to introduce the \textbf{descendant cascade} of a t-object $\bt^\alpha_\beta$, which we define as a set of t-objects comprising $\bt^\alpha_\beta$ and all of its descendants. For $\alpha,\beta,\mu,\nu\in\aleph$, we denote the descendant cascade of $\bt^\alpha_\beta$ as
\[
    D(\bt^\alpha_\beta) = \{\bt^\mu_\nu: \mu\cap\alpha=\alpha,\; \nu\cap\beta=\beta,\; \mu\cap\nu=\varnothing\}
\]

\subsection{Obliqueness}\label{subsec:obliqueness}

The decomposition of t-objects establishes a method for identifying their linear dependence and obliqueness. We say that two nonidentical t-objects are \textbf{oblique} if their descendant cascades contain coinciding t-objects.
The following lemma establishes that our definition of orthogonality is equivalent to the property of not being oblique.

\textbf{Lemma}. T-objects are nonoblique if and only if they are orthogonal.
\begin{proof}
    Suppose $\bt^{\alpha_1}_{\beta_1} \perp \bt^{\alpha_2}_{\beta_2}$. This implies that either $\alpha_1\cap \beta_2\not =\varnothing$ or $\alpha_2\cap \beta_1\not =\varnothing$ (or both). Let us consider the case where $\alpha_1\cap \beta_2\not =\varnothing$. Then there must be an index $i$ which is element of both
    \[
        i\in \alpha_1, \beta_2
    \]
    Hence, all descendants of $\bt^{\alpha_1}_{\beta_1}$ will have $i$ as an upper index, and all descendants of $\bt^{\alpha_2}_{\beta_2}$ will have $i$ as a lower index; hence, there will be no identical descendants.

    Now, suppose that the two t-objects $\bt^{\alpha_1}_{\beta_1}$ and $\bt^{\alpha_2}_{\beta_2}$ are not oblique. Then, they must be orthogonal. Indeed, let us consider the product
    \[
        \bt^{\alpha_1}_{\beta_1}\,\bt^{\alpha_2}_{\beta_2} = \bt^{\alpha_1\,\alpha_2}_{\beta_1\,\beta_2}
    \]
    If this product is not empty, it is obviously a descendant of both $\bt^{\alpha_1}_{\beta_1}$ and $\bt^{\alpha_2}_{\beta_2}$, which contradicts the assumption.
\end{proof}

Note that the obliqueness of t-objects makes the task of identifying the equality of state vectors rather nontrivial in general. Here is an example of a vanishing state vector:
\[
    \bs = \bt_{123} + \bt^{123} + \bt^2_1 + \bt^3_2 + \bt^1_3 - \bt = 0
\]
It is possible to verify that this state vector vanishes by expanding all t-objects by indices~1, 2, and 3. This example has already been encountered in Eq.~\eqref{eq:trivial_irreducible}.

\subsection{Orthogonalisation}\label{subsec:orthogonalisation}

Consider two arbitrary t-objects
\begin{align*}
    & \bt_{(1)} = \bt^{\alpha_1}_{\beta_1} \\
    & \bt_{(2)} = \bt^{\alpha_2}_{\beta_2}
\end{align*}
The t-object $\bt_{(1)}$ can be decomposed into two components with respect to $\bt_{(2)}$: a collinear component $\bT_{\parallel}$ and an orthogonal component $\bT_{\perp}$:
\[
    \bt_{(1)} = \bT_{\parallel} + \bT_{\perp}
\]
We define these components as follows. Let us designate
\[
    \gamma = \alpha_2 \smallsetminus \alpha_1\,,\qquad \nu = \beta_2 \smallsetminus \beta_1
\]
For the sake of demonstration, let's assume that the sets $\gamma$ and $\nu$ each contain two indices
\[
    \gamma = \{ij\}\,,\quad \nu = \{lm\}
\]
Generalisation to any other combination of indices would be straightforward. It is easy to see that the vector $\bt_{(1)}$ can be represented as
\begin{align*}
    & \bt^{\alpha_1}_{\beta_1} =
    \bt^{\alpha_1\,ij}_{\beta_1\,lm} +
    \bt^{\alpha_1\,i}_{\beta_1\,lmj} +
    \bt^{\alpha_1}_{\beta_1\,lmi} +
    \bt^{\alpha_1\,m}_{\beta_1\,l} +
    \bt^{\alpha_1\,l}_{\beta_1}
\end{align*}
To see why this identity holds, we can iteratively apply atomic reduction to the leftmost terms of the expression on the right-hand side. For instance, the first two terms can be reduced by index $j$ to yield
\[
    \bt^{\alpha_1\,ij}_{\beta_1\,lm} + \bt^{\alpha_1\,i}_{\beta_1\,lmj} = \bt^{\alpha_1\,i}_{\beta_1\,lm}
\]
The result of the first reduction can be reduced with the next term by eliminating index $i$, and so on, until all indices $\{i, j, k, l \}$ are eliminated.

The first term in this series is the collinear component, defined as the product of two t-objects:
\begin{align*}
    &\bT_{\parallel} = \bt_{(1)}\,\bt_{(2)} = \bt^{\alpha_1\,\alpha_2}_{\beta_1\,\beta_2} =
    \bt^{\alpha_1\,ij}_{\beta_1\,lm}
\end{align*}
The rest of the series is the orthogonal component
\begin{align*}
    &\bT_{\perp} = \bt^{\alpha_1\,i}_{\beta_1\,lmj} +
    \bt^{\alpha_1}_{\beta_1\,lmi} +
    \bt^{\alpha_1\,m}_{\beta_1\,l} +
    \bt^{\alpha_1\,l}_{\beta_1}
\end{align*}
such that
\[
    \bt_{(1)} = \bT_{\parallel} + \bT_{\perp}
\]

Let us prove some properties of the collinear and orthogonal components. First, we notice that both are binary and mutually orthogonal:
\begin{align*}
    & \bT_{\parallel}\,\bT_{\parallel} = \bT_{\parallel} \\
    & \bT_{\perp}\,\bT_{\perp} = \bT_{\perp} \\
    & \bT_{\parallel}\, \bT_{\perp} = 0
\end{align*}
The binary nature of $\bT_{\perp}$ is a direct consequence of its construction as the sum of pairwise orthogonal t-objects. Next, we check that the orthogonal component is indeed orthogonal to~$\bt_{(2)}$:
\begin{align*}
    & \bt_{(2)}\,\bT_{\perp} = 0
\end{align*}
This can be easily verified by direct substitution.

\subsection{Finite Dimensional Space} \label{subsec:finite-dimensional-space}

In this section, we consider $N < \infty$, such that the space of all t-objects is finite.

\subsubsection{Pure States}

For a finite $N$, a \textbf{pure state} is defined as a nonzero t-object for which the upper and lower indices cover the entire~$\mathbb{N}$:
\begin{align}
    \bt^\alpha_\beta\,, \qquad \alpha \cup\beta = \mathbb{N}\,, \quad \alpha\cap\beta=\varnothing
\end{align}
There are $2^N$ distinct pure states. They can be obtained by all permutations of the indices between upper and lower positions. For example, for $N=3$ we have the following eight pure states:
\[
    \bt^{123}_{}, \quad
    \bt^{12}_{3}, \quad
    \bt^{13}_{2}, \quad
    \bt^{23}_{1}, \quad
    \bt^{1}_{23}, \quad
    \bt^{2}_{13}, \quad
    \bt^{3}_{12}, \quad
    \bt^{}_{123}
\]

Pure states form the orthogonal basis of $\mathbb{S}$ owing to the following two key properties:
\begin{itemize}
    \item \textbf{Mutual Orthogonality}: Pure states are orthogonal to each other. This is obvious because they have no descendants, which means they are not oblique, and hence, they are orthogonal according to the lemma.
    \item \textbf{Decomposability}: Any t-object can be expressed as a sum of pure states. Indeed, suppose $\bt^\alpha_\beta$ is not a pure state. Then, there are some indices that are not present in either $\alpha$ or $\beta$. Let us designate the set of those indices $\gamma$:
    \[
        \gamma = \mathbb{N} \smallsetminus (\alpha \cup \beta)
    \]
    Suppose $\gamma$ contains $n$ indices: $\gamma = \{i_1, \ldots,i_n\}$. At first step we can use decomposition by index $i_1$ to get a sum of two t-objects
    \[
        \bt^\alpha_\beta = \bt^{\alpha\,i_1}_\beta + \bt^\alpha_{\beta\,i_1}
    \]
    At the next step we apply decomposition by index $i_2$ to both terms to obtain
    \[
        \bt^\alpha_\beta = \bt^{\alpha\,i_1\,i_2}_\beta + \bt^{\alpha\,i_1}_{\beta\,i_2} +
        \bt^{\alpha\,i_2}_{\beta\,i_1} + \bt^{\alpha}_{\beta\,i_1\,i_2}
    \]
    The decomposition process continues until all $n$ indices $\{i_1,\ldots,i_n\}$ are added to each term. The final decomposition contains $2^n$ terms obtained by all permutations of the indices $\{i_1,\ldots,i_n\}$ between the upper and lower positions. For instance, for $N=3$,
    \[
        \bt^1 = \bt^{123}_{} + \bt^{12}_{3} + \bt^{13}_{2} + \bt^{1}_{23}
    \]
\end{itemize}
Consequently, any state vector can be represented as a linear combination of the pure states. Thus, we conclude that the dimension of $\mathbb{S}$ in the case of finite $N$ is $2^N$.

\subsection{Infinite Dimensional Space}\label{subsec:infinite-dimensional-space}

Although this study does not develop a theory of infinite-dimensional algebra of t-objects, we would like to highlight two new aspects that emerge when $N$ is allowed to be infinite.

Firstly, since we defined state vectors as finite linear combinations
\[
    \bs = \sum_{i = 1}^k a_i\,\bt^{\alpha_i}_{\beta_i}\,,\qquad
    \alpha_i,\beta_i \in\aleph\,,\qquad a_i\in\mathbb{Z}\,,\qquad k<\infty
\]
space $\mathbb{S}$ becomes open. The closure $\bar{\mathbb{S}}$ contains the limits of all sequences
\[
    \lim_{k\to\infty} \sum_{i = 1}^k a_i\,\bt^{\alpha_i}_{\beta_i}
\]

Our second point addresses the basis. Given that the index sets of a t-object must be finite, introducing pure states for infinite $N$ is impossible. Indeed, a pure state is defined as $\bt^\alpha_\beta$ such that $\alpha\cup\beta=\mathbb{N}$. However, if $N$ is infinite, so is $\mathbb{N}$, and hence, at least one of $\alpha$ or $\beta$ must be infinite.

Pure states can be introduced as limits of the products. For instance:
\[
    \bt^{123\ldots} = \bt^1\,\bt^2\,\bt^3\ldots = \prod_{i=1}^\infty \bt^i
\]
They can also be obtained as limits of the series. For instance:
\[
    \bt^{123\ldots} = \bt - \bt_1 - \bt^1_2 - \bt^{12}_3 - \bt^{123}_4 - \ldots
\]

For a finite set of t-objects $\{\bt^{\alpha_1}_{\beta_1}, \ldots , \bt^{\alpha_n}_{\beta_n}\}$ it is always possible to introduce a local basis. Suppose $I$ is the set of all indices covered by the given set of t-objects:
\[
    I = \alpha_1 \cup  \ldots \cup \alpha_n \cup \beta_1 \cup \ldots \cup \beta_n =
    \{i_1, \ldots,i_m\}
\]
We can introduce a finite set of \textbf{constrained pure states} for the set of indices $I$, which consists of t-objects
\[
    \{\bt^\alpha_\beta\,:\quad \alpha\cap\beta=\varnothing\,,\quad \alpha\cup\beta = I\}
\]
Constrained pure states constitute an orthogonal basis in the space defined by indices $I$.

\section{Algorithms of Logical Inference}\label{sec:algorithms}

In this section, we explore logical inference algorithms using state algebraic approach. We begin by examining the computer representation of state-algebraic objects, as this influences the computational algorithms. Subsequently, we detail the algorithms for common inference tasks and analyse their complexity.

In this study, we do not delve deeply into the topic of optimisation heuristics, as it is an area of active research that deserves a separate discussion. The algorithms presented herein consider the most direct brute force implementations. Consequently, the estimated complexity of the algorithms reflects the worst-case scenarios without considering any optimisation possibilities. We provide only a brief overview of some optimisation opportunities specific to State Algebra in Section~\ref{subsec:optimisation}.

\subsection{Storing State-Algebraic Objects}\label{subsec:storing-s-algebraic-objects}

There are many different ways in which a state vector can be represented in computer programs. In this paper, we consider the state vectors in the row decomposition. Recall that raw decomposition represents a state vector as a sum of t-objects
\[
    \bs = \bt^{\alpha_1}_{\beta_1} + \bt^{\alpha_2}_{\beta_2} +
    \ldots + \bt^{\alpha_k}_{\beta_k}
\]
In this representation, the state vector can be stored in the memory as a list of t-objects.

There is a simple natural way to represent a t-object $\bt^\alpha_\beta$ in a computer program as a combination of lists of upper and lower indices $(\alpha, \beta)$.

\subsection{Calculation of Valid Set}\label{subsec:calculation-of-valid-set}

Every logical formula in State Algebra can be represented by the corresponding state vector, which in turn can be decomposed into a sum of t-objects
\[
    f\sim\bs \qquad\text{where }\bs = \bt^{\alpha_1}_{\beta_1} + \bt^{\alpha_2}_{\beta_2} + \ldots
\]
Note that an arbitrary formula as a function of $n$ variables can be represented as the sum of no more than $2^{n-1}$ t-objects.\footnote{This is easy to prove by induction. Indeed, the statement is correct for a formula of one variable, since t-objects $\bt$, $\bt_i$ and $\bt^i$ (where $i$ is the index of the variable) exhaust all the possibilities. Thus, if $n=1$, then any formula can be expressed as $2^0=1$ row. Let for an arbitrary $n$, some formula has a corresponding state vector~$\bs$. For an arbitrary variable with index $j$, the state vector can be split as $\bs = \bs^j + \bs_j$, where $\bs^j$ and $\bs_j$ represent the state vectors corresponding to some formulas of $n-1$ variables. If $\bs^j$ and $\bs_j$ can be expressed as no more than $2^{n-2}$ rows, then $\bs$ has no more than $2^{n-1}$ rows.}

Suppose we are given a set of state vectors representing the premise, and we would like to calculate the complete valid set.

First, we consider the product of two state vectors. Let
\begin{align*}
    &\bs_{(1)} = \bt_{(1)(1)} + \bt_{(1)(2)} + \ldots + \bt_{(1)(n_1)} \\
    & \bs_{(2)} = \bt_{(2)(1)} + \bt_{(2)(2)} + \ldots + \bt_{(2)(n_2)}
\end{align*}
The product $\bs_{(1)}\,\bs_{(2)}$ contains $n_1\,n_2$ pair-wise products of t-objects:
\begin{align*}
    \bs_{(1)}\,\bs_{(2)} = \sum_{i=1}^{n_1} \sum_{j=1}^{n_2} \bt_{(1)(i)}\,\bt_{(2)(j)}
\end{align*}
Thus, the complexity of this operation is $O(n_1\,n_2)$.

Each product of t-objects results into a single t-object according to the rule
\[
    \bt^{\alpha_1}_{\beta_1}\, \bt^{\alpha_2}_{\beta_2} = \bt^{\alpha_1\,\alpha_2}_{\beta_1\,\beta_2}
\]
which means that the algorithm only needs to combine the lists of the upper and lower indices for each pair of t-objects. Note that the product of t-objects vanishes if the resulting sets of upper and lower indices overlap; hence, the number of t-objects in the product $\bs_{(1)}\,\bs_{(2)}$ may be smaller than~$n_1\,n_2$.

Suppose now, the knowledge base consists of $m$ state vectors
\begin{align*}
    & \bs_{(1)} = \bt_{(1)(1)} + \bt_{(1)(2)} + \ldots + \bt_{(1)(n_1)} = \sum_{j=1}^{n_1} \bt_{(1)(j)} \\
    & \bs_{(2)} = \bt_{(2)(1)} + \bt_{(2)(2)} + \ldots + \bt_{(2)(n_2)} = \sum_{j=1}^{n_2} \bt_{(2)(j)} \\
    & \vdots \\
    & \bs_{(m)} = \bt_{(m)(1)} + \bt_{(m)(2)} + \ldots + \bt_{(m)(n_m)} = \sum_{j=1}^{n_m} \bt_{(m)(j)}
\end{align*}
where $\bt_{(i)(j)}$ is the $j$-th row of the $i$-th state vector, and $i$-th state vector has $n_i$ rows. The complete valid set is given as a product of all state vectors
\begin{align*}
    \bV = \bs_{(1)}\,\bs_{(2)}\ldots \bs_{(m)}
\end{align*}
We can compute the product by first computing $\bs_{(1)}\,\bs_{(2)}$, then multiplying the result by $\bs_{(3)}$, and so on, until the last vector, $\bs_{(m)}$. In the worst-case scenario, if none of the t-objects vanishes, the total number of pair-wise t-object multiplications, and hence, the worst-case complexity of the algorithm will be $O(n_1\,n_2\ldots n_m)$.

Suppose that all state vectors in the knowledge base have at most $K$ t-objects. Then, the estimate of the upper limit of the algorithm complexity is $O(K^m)$, where $m$ is the number of logical formulas in the premise (size of the knowledge base).

The resulting valid set consists of up to $K^m$ t-objects. Practically, this number is often smaller; some t-objects vanish, and the remaining t-objects can undergo a reduction procedure, which can further reduce the total number of t-objects.

In Appendix~\ref{sec:example-of-logical-inference} we demonstrate a toy example of how the computation of a valid set and logical inference can be performed using state algebraic operations.

\subsection{Simplifying Expressions with T-objects}\label{subsec:simplifying-expressions-with-t-objects}

A sum of t-objects can often be transformed into an equivalent but simpler expression using the following steps:
\begin{enumerate}
    \item Identifying and removing vanishing t-objects:
    \[
        \bt^{\alpha}_{\beta} = 0 \quad \text{if} \quad \alpha\cap\beta \not = \varnothing
    \]
    \item Using atomic reduction to reduce the number of t-objects. Atomic reduction can be applied recursively in many different ways using strategies driven by optimisation heuristics. In particular, depending on the task to be solved, one can choose to apply canonical or noncanonical reduction.
    \item If state vectors are not binary, the expression with t-objects resulting from multiplication can contain some redundancy. Redundant t-object can be removed by using the identity from Eq.~\eqref{eq:reducdent_states}
    \begin{align*}
        & [\bt^{\alpha\,\mu}_{\beta\,\nu} + \bt^{\alpha}_{\beta}] = \bt^{\alpha}_{\beta}\qquad
        \forall \mu\,,\nu \in\aleph
    \end{align*}
    The latter step does not preserve multiplicity factors. It can be performed only if one is interested in the result of the inference process up to an equivalence transformation.
\end{enumerate}

\subsection{Reduction}\label{subsec:reduction-of-state-vectors}

Suppose that we have a list of $K$ t-objects. The brute-force reduction algorithm repeats iterations, where one iteration goes into a nested loop over $K\,(K-1)/2\approx K^2$ distinct pairs of t-objects to apply atomic reductions. To estimate the worst-case scenario, suppose that in one iteration, we identify exactly $K/2$ pairs of reducible t-objects, such that at the end of the iteration, the state vector consists of $K/2$ new t-objects. Because all the remaining t-objects are modified, we need to perform the same iteration again, but this time for $K/2$ objects. Repeating this procedure recursively, the algorithm needs to perform the order of magnitude of
\[
    K^2 + (K/2)^2 + (K/4)^2 + \ldots < \frac{4}{3} K^2
\]
pair-wise operations. Thus, we estimate the upper limit of the complexity of the full reduction algorithm to be $O(K^2)$.

In practice, the algorithm usually terminates earlier when no reducible pairs of t-objects remain.

The algorithm above considers a brute force noncanonical procedure. Optimisation heuristics can be applied to achieve a ``good enough'' level of compression in a much shorter time.

If the canonical structure must be preserved during reduction, a modified reduction procedure should be applied, as described in the following section.

\subsubsection{Preserving Canonical Structure}\label{subsubsec:preserving-canonisity}

Even if the operands of the state vector multiplication are in the canonical form, multiplication may result in a noncanonical state vector. For instance:\footnote{In this section we will use the subscript $C$ to indicate that a state vector is either in a canonical form, or can be reduced to the canonical form by ordered reduction.}
\[
    \begin{Bmatrix}
        - & 1 & - \\
        1 & 0 & 1
    \end{Bmatrix}_C \cdot
    \begin{Bmatrix}
        - & - & 1
    \end{Bmatrix}_C =
    \begin{Bmatrix}
        - & 1 & 1 \\
        1 & 0 & 1
    \end{Bmatrix}
\]
In this example, both state vectors on the left-hand side are reduced using an ordered procedure. Indeed, the first operand is reduced by index 3, followed by index 1, as shown below:
\[
    \begin{Bmatrix}
        1 & 1 & 1 \\
        1 & 1 & 0 \\
        0 & 1 & 1 \\
        0 & 1 & 0 \\
        1 & 0 & 1
    \end{Bmatrix}_C =
    \begin{Bmatrix}
        1 & 1 & - \\
        0 & 1 & - \\
        1 & 0 & 1
    \end{Bmatrix}_C =
    \begin{Bmatrix}
        - & 1 & - \\
        1 & 0 & 1
    \end{Bmatrix}_C
\]
The canonicity of the second operand is evident. The product however is not in a canonical form, since, if fully expanded and then reduced again by an ordered procedure, we would obtain
\[
    \begin{Bmatrix}
        1 & - & 1 \\
        0 & 1 & 1
    \end{Bmatrix}_C
\]
To preserve the canonical form of the product, the atomic reduction procedure must be modified as follows. We say that a pair of rows is left-reducible by index $i$ if they are reducible directly by index $i$, or if any of their left descendants are reducible directly. Here, the left descendants of a row are t-objects obtained by expanding the row by indices $j < i$. If the rows are left-reducible, they should be expanded by the indices $j < i$, after which the reduction procedure is continued.

In the example above, we note that the two rows $\{-\ 1\  1\}$ and $\{1\ 0\  1\}$ can be reduced by index 2 if the first row is expanded by index 1. Therefore, we proceed as follows:
\[
    \begin{Bmatrix}
        - & 1 & 1 \\
        1 & 0 & 1
    \end{Bmatrix} =
    \begin{Bmatrix}
        1 & 1 & 1 \\
        0 & 1 & 1 \\
        1 & 0 & 1
    \end{Bmatrix}_C =
    \begin{Bmatrix}
        1 & - & 1 \\
        0 & 1 & 1
    \end{Bmatrix}_C
\]
We see that there is a price to pay for preserving the canonical form under multiplication. In some situations, the updated reduction procedure may lead to a significant increase in the size of intermediate state vectors. Below, we discuss the trade-off between the canonical and noncanonical approaches.

\subsection{Subtraction}\label{subsec:subtraction-of-state-vectors}

The coordinate-wise subtraction of state vectors is not defined in the row decomposition. We are interested in the subtraction of state vectors only in a set-theoretical sense. Namely, as a difference between $\bs_{(1)}$ and $\bs_{(2)}$ we want to find a state vector, which, up to equivalence transformation, contains all states from $\bs_{(1)}$ that are not present in $\bs_{(2)}$. We use the set difference notation $\smallsetminus$ for this operation.

First, we consider the subtraction of a t-object from the state vector. Let
\begin{align*}
    & \bs = \bt_{(1)} + \bt_{(2)} + \ldots + \bt_{(n)}
\end{align*}
We consider the following expression:
\[
    \bs \smallsetminus \bt^{\alpha}_{\beta}
\]
To ensure that the states contained in  $\bt^{\alpha}_{\beta}$ are removed from $\bs$, we must remove those from each t-object:
\begin{align*}
    \bs \smallsetminus \bt^{\alpha}_{\beta}& = (\bt_{(1)}\smallsetminus \bt^{\alpha}_{\beta})  + (\bt_{(2)}\smallsetminus \bt^{\alpha}_{\beta}) + \ldots \\
    & = [\bt_{(1)} -\bt^{\alpha}_{\beta}]  + [\bt_{(2)} - \bt^{\alpha}_{\beta}] + \ldots
\end{align*}
Now we can use Eq.~\eqref{eq:subtract_t_t} to perform calculations.

The subtraction of the two state vectors is an obvious extension of the previous step. Let
\begin{align*}
    &\bs_{(1)} = \bt_{(1)(1)} + \bt_{(1)(2)} + \ldots + \bt_{(1)(n_1)} \\
    & \bs_{(2)} = \bt_{(2)(1)} + \bt_{(2)(2)} + \ldots + \bt_{(2)(n_2)}
\end{align*}
Subtraction can be performed iteratively as follows:
\begin{align}
    \bs_{(1)} \smallsetminus \bs_{(2)}& =
    \left(\left(\left(\bs_{(1)}\smallsetminus \bt_{(2)(1)}\right)
              \smallsetminus \bt_{(2)(2)}\right) \smallsetminus \cdots
    \right)\smallsetminus \bt_{(2)(n_2)}
\end{align}
The complexity of the state vector subtraction can be estimated as follows: Let $K$ be the upper bound on the number of rows in the state vectors, and $L$ be the maximum number of indices in any t-object (equivalent to the number of \bound columns in the state vectors).

According to Eq.~\eqref{eq:subtract_t_t}, subtracting two t-objects yields no more than $L$ new t-objects. Therefore, in the worst case, subtracting a t-object from a state vector requires $K$ pairwise t-object subtractions, resulting in $K \cdot L$ new t-objects being created. Consequently, recursively applying subtraction to two state vectors will require approximately
\[
    O(K + L\,K + L^2\,K + \ldots + L^{K-1}\,K)\approx O(K\cdot L^{K-1})
\]
pair-wise t-object subtraction operations, and will result in $O(K\cdot L^K)$ new t-objects.

In certain scenarios, an equivalent operation may offer a performance advantage:
\[
    \bs_{(1)} \smallsetminus \bs_{(2)} \cong \bs_{(1)} \smallsetminus (\bs_{(1)}\,\bs_{(2)})
\]
This is particularly true if the product $\bs_{(1)}\,\bs_{(2)}$ contains significantly fewer t-objects than the vector~$\bs_{(2)}$.

\subsection{Asserting State Vectors Equivalence}\label{subsec:asserting-state-vectors-equality}

Suppose that there are two state vectors $\bs$ and $\bq$, and we need to assert that they are equal up to the equivalence transformation:
\[
    \bs \cong \bq
\]

If the state vectors are not in a canonical form, this assertion cannot be easily implemented by comparing the state vectors row-by-row because even if the state vectors have the same coordinates, they can have different equivalent representations. To verify this equality, we can rewrite the assertion using a set-theoretical identity:
\[
    \bs \cong \bq \qquad \iff \qquad
    \bs \smallsetminus \bq = 0 \quad \land\quad \bq\smallsetminus \bs = 0
\]
The latter identities can be asserted because, as noted before, in row decomposition, an empty state vector is literally empty. There exists no equivalent representation of the empty state vector other than that with no rows.

If one of the vectors is trivial, the condition simplifies to
\[
    \bs \cong \bt \qquad \iff \qquad \bt \smallsetminus \bs = 0
\]

\subsection{Optimisation}\label{subsec:optimisation}

We would like to start by re-emphasising the point made in the introduction: State Algebra is a generic framework for the computation and abstract representation of mathematical logic. Its practical implementation offers diverse optimisation possibilities. The effectiveness of state-algebraic computational algorithms, compared to other techniques, largely depends on the specific problem and chosen optimisation heuristics. For instance, while ROBDDs can achieve a more compact representation of logical formulas,\footnote{
    For every binary decision diagram, we can find a corresponding state vector such that the rows represent paths on the BDD, whereas the holes signify the eliminated nodes in an ROBDD. The recombining tree topology of the ROBDD, as a rule, leads to a higher number of paths than nodes.
} State Algebra's flexibility in noncanonical reduction can sometimes lead to more efficient representations.

State Algebra also necessitates the implementation of a different set of operations. Instead of graph operations, most state-algebraic operations are reduced to repeated pairwise multiplications of t-objects. This approach facilitates relatively straightforward implementation algorithms and allows the parallelisation of most operations.

This paper does not extensively cover optimisation algorithms, as this is a topic for separate research. Instead, we only highlight some promising avenues for optimisation exploration.

\subsubsection{Reordering Heuristics}

The efficiency of the algorithms for both operations -- reduction and multiplication of state vectors -- may strongly depend on the order in which the operations are performed. For instance, the reduction of two t-objects is only possible if all but one of their indices are identical. This can be used by a heuristic optimisation of the reduction algorithm: by changing the order of atomic reductions, it is possible to increase the probability of successful matching of reducible pairs of t-objects.

Similarly, for multiplication, when a large number of state vectors are multiplied, the size of the intermediate state vector strongly depends on the order in which the vectors are multiplied. Products containing more mutually orthogonal t-objects result in smaller intermediate-state vectors. Therefore, an optimisation algorithm can reorder and group state vectors that have a higher chance of ``colliding'' t-objects. This can be achieved, for instance, by clustering state vectors according to their similarity in the space of pivot sets.

\subsubsection{Canonical Versus Noncanonical Representation}

As mentioned previously, when performing a reduction operation, one can choose between ordered and unordered versions. Ordered reduction requires fixing the order of variables and performing all atomic reductions in a predetermined order of index elimination. Conversely, for noncanonical reduction, one has the freedom to choose the order in which atomic reduction is applied. There may be different reasons for preferring either the ordered or unordered approach.
\begin{itemize}
    \item If reduced by an ordered procedure (i.e., in canonical form), state vectors have unique representation in row decomposition. This allows a polynomial-time comparison of the two state vectors. A well-designed code implementation can further optimise this to a constant-time pointer comparison.
    \item The canonical form requires fixing the variable order for all formulas in the knowledge base. This ordering can significantly affect the compression achieved by reduction. An order optimal for one formula may not be optimal for another, potentially leading to a substantial increase in the size of the state vectors during intermediate logical inference steps.
    \item If the inference task does not require state vector comparison, better optimisation may be achieved by using a bespoke heuristic optimisation for each state vector that needs to be reduced, without relying on a predefined ordering. This offers additional flexibility when designing an optimisation for a specific inference task.
\end{itemize}

\subsubsection{Equivalent Formulations of Inference Objective}

The goal of inference can often be achieved through various methods, with the optimal choice dictated by the optimisation. Consider an example in which we have a knowledge base of a number of formulas and aim to determine whether a specific relationship, expressed as a target logical formula $f$, holds for the valid set. If $f \sim \bq$ represents the target state vector, the following approaches can be used to answer this question (completely or partially):

\begin{enumerate}
    \item Create an extended formula using an indicator event $E_k=f$ and add a corresponding extended state vector to the knowledge base. The value of $E_k$ should then be checked after multiplying all the state vectors.
    \item Find the intersection between the valid set $\bV$ and the information vector $\hat \bq = \bt\smallsetminus \bq$
    \[
        \bV \,\hat\bq
    \]
    If this intersection is empty, the possibility of the relationship being false is ruled out. However, this method does not exclude the possibility that the relationship is contradictory in valid space.
    \item Find an intersection between the information set $\hat \bI$ and the vector $\hat \bq$:
    \[
        \hat \bI\,\hat \bq
    \]
    If the cardinality of this product is less than the cardinality of $\hat \bq$
    \[|\hat \bI\,\hat \bq| < |\hat \bq| \]
    then the valid set will inevitably contain states from $\hat\bq$, and thus, $f$ will be falsified for those states within the valid set. To prove this inequality, it is sufficient to find at least one state in $\hat \bq$ that is incompatible with $\hat \bI$.
\end{enumerate}
The choice among these equivalent approaches depends on the specific problem and its potential for optimisation.

Note that to prove that the intersection of a number of state vectors is not empty, it suffices to find at least one combination of corresponding t-objects with a nonvanishing product. Conversely, to prove that the product of the state vectors is empty, it is sufficient to find a minimal incompatible subset of all the state vectors. Heuristic \emph{search-based} optimisation methods can be applied to these types of problems.

Algorithm optimisation may also vary based on whether a computation is performed once or multiple times.

When a complete valid set must be computed before an inference problem can be solved, the process falls under the category of \emph{knowledge compilation} algorithms. A practical example is rule-based classification, such as in credit card fraud detection. A predefined set of rules constitutes the knowledge base. Each transaction possesses attributes that serve as evidence and can be expressed as a single t-object. The transaction is then checked against the knowledge base to determine the value of a target variable (e.g. ``fraud''). There are four possible outcomes: ``fraud'' is true, false, indefinite, or the evidence contradicts the knowledge base (in which case meta-rules typically guide decision-making). In scenarios involving millions of transactions classified by the same knowledge base, pre-calculating the valid set can significantly accelerate the real-time transaction classification.

However, if classification is a one-time operation, it may be more efficient to add evidence directly to the knowledge base and then optimise the computation of the extended knowledge base.

\section{Conclusion and Extensions}\label{sec:enhancements-and-future-work}

This paper introduced State Algebra, a framework that recasts problems in propositional logic into an algebraic system. A central design goal was to achieve both efficiency and flexibility. Following the rigorous and consistent development of the theory, we introduced a layered architecture comprising Set, Coordinate, and Row Decomposition representations. This structure motivates the algebra of T-objects -- an abstract-algebraic formulation of propositional logic, whose fundamental mathematical properties we have outlined, without being exhaustive.

The State Algebra framework can be naturally extended to probabilistic logic, also known as Markov Random Fields. By replacing the integer multiplicity factors in the coordinate representation with real-valued factors, State Algebra is transformed into a vector space spanned by t-objects and equipped with a multiplication operation. In this extended framework, the state vectors represent probability distributions over the state space. The algebraic operations within this system directly map to operations on probabilities: multiplication of state vectors is equivalent to factoring a joined distribution, whereas the summation of coordinates corresponds to marginalisation. This extension allows State Algebra to function as a framework for representing Markov Random Fields, thereby providing a tool for Weighted Model Counting (WMC), a central problem in probabilistic inference.

This foundation enables the development of sophisticated applications, such as accurate and interpretable decisioning systems with broad applicability\footnote{For a specific example in the context of healthcare we refer to~\cite{dhaese-et-al:2021}.}. The presentation of these results is beyond the scope of this study and will be published separately.

The second extension of State Algebra is its application to higher-order logic (HOL). In comparison with propositional logic, HOL is a much more complex construct, where events are not the primary building blocks but emerge from applying predicates to objects of an abstract nature. Logical sentences in HOL can operate with these objects and use quantifiers. The concept of logical inference in HOL is significantly different. Similar to how the relativistic extension of quantum mechanics addresses the dynamic creation and annihilation of particles, the transition to HOL requires the handling of dynamically generated events and ground formulas during logical inference. By incorporating mechanisms to manage this dynamic universe, State Algebra can be transformed into an effective tool for expressing logical sentences and performing inferences in HOL\@. The detailed development of the theories underpinning these extensions lies beyond the scope of this paper and will be the subject of future publications.

\pagebreak

\addcontentsline{toc}{section}{Appendix}

\begin{Large}
    \textbf{Appendix}
\end{Large}

\appendix

\section{Example of Logical Inference}\label{sec:example-of-logical-inference}

Let us consider an example of the ``Importation'' rule in propositional logic:
\[
    (E_1 \to (E_2 \to E_3)) \vdash ((E_1 \land E_2) \to E_3)
\]
where $E_1, E_2, E_3$ are propositional variables.

We define a space of eight events $\{E_1, \ldots,E_8\}$ and introduce the following logical formulas:
\begin{align}
    \label{eq:E4}
    & E_4 = ( E_2 \to E_3)\\
    \label{eq:E5}
    &E_5 = (E_1 \to E_4) \\
    \label{eq:E6}
    & E_6 = (E_1 \land E_2)\\
    \label{eq:E7}
    &E_7 = (E_6 \to E_3)
\end{align}
Note that events $E_4, E_5, E_6, E_7$ are supplementary. We want to prove that if the premise is true, then the consequent is also true. Therefore we add another formula
\begin{align}
    \label{eq:E8}
    & E_8 = (E_5 \to E_7)
\end{align}
where event $E_8$ is an indicator event of the formula that we want to prove. The goal of logical inference is to show that $E_8$ is identically true.

The state vectors corresponding to the relations (\ref{eq:E4} -~\ref{eq:E8}) are
\begin{align}
    & (\ref{eq:E4}):\qquad
    \begin{Bmatrix}
        -  &  1  &  1   &  1  &  -  & -  & - & - \\
        -  &  0  &  -   &  1  &  -  & -  & - & - \\
        -  &  1  &  0   &  0  &  -  & -  & - & -
    \end{Bmatrix} =
    \bt^{234} + \bt^4_2 + \bt^2_{34} \\
    & (\ref{eq:E5}):\qquad
    \begin{Bmatrix}
        1  &  -  &  -  &  1  &  1  & -  & - & - \\
        0  &  -  &  -  &  -  &  1  & -  & - & - \\
        1  &  -  &  -  &  0  &  0  & -  & - & -
    \end{Bmatrix} =
    \bt^{145} + \bt^5_1 + \bt^1_{45} \\
    & (\ref{eq:E6}):\qquad
    \begin{Bmatrix}
        1  &  1  &  -  &  -  &  -  & 1  & - & - \\
        1  &  0  &  -  &  -  &  -  & 0  & - & - \\
        0  &  -  &  -  &  -  &  -  & 0  & - & -
    \end{Bmatrix} =
    \bt^{126} + \bt^1_{26} + \bt_{16} \\
    & (\ref{eq:E7}):\qquad
    \begin{Bmatrix}
        -  &  -  &  1  &  -  & - & 1  & 1 & - \\
        -  &  -  &  -  &  -  & - & 0  & 1 & - \\
        -  &  -  &  0  &  -  & - & 1  & 0 & -
    \end{Bmatrix} =
    \bt^{367} + \bt^7_6 + \bt^6_{37} \\
    & (\ref{eq:E8}):\qquad
    \begin{Bmatrix}
        -  &  -  &  -  &  -  &  1  &  - & 1  & 1 \\
        -  &  -  &  -  &  -  &  0  &  - & -  & 1 \\
        -  &  -  &  -  &  -  &  1  &  - & 0  & 0
    \end{Bmatrix} =
    \bt^{578} + \bt^8_5 + \bt^5_{78}
\end{align}

To calculate the valid set, we multiply the state vectors with each other one-by-one, as described above. As an illustration, let us consider the product of the first two state vectors:
\[
    (\bt^{234} + \bt^4_2 + \bt^2_{34}) (\bt^{145} + \bt^5_1 + \bt^1_{45})
\]
Of the nine terms resulting from the multiplication, three vanish because of coinciding indices. Atomic reduction can be applied to the remaining terms to simplify the expression. The result is as follows (repeating indices can be ignored):
\begin{align}
    & (\bt^{234} + \bt^4_2 + \bt^2_{34}) (\bt^{145} + \bt^5_1 + \bt^1_{45}) = \\
    & \qquad = \bt^{234145} + \bt^{2345}_{1} + \bt^{2341}_{45}  + \bt^{4145}_{2} + \bt^{45}_{21} + \bt^{41}_{245} +
    \bt^{2145}_{34} + \bt^{25}_{341} + \bt^{21}_{3445} \\
    \label{eq:5.15}
    & \qquad = \bt^{23415} + \bt^{2345}_{1} + \bt^{415}_{2} + \bt^{45}_{21} + \bt^{25}_{341} + \bt^{21}_{345} \\
    \label{eq:5.16}
    &\qquad  = \bt^{2345} + \bt^{45}_{2} + \bt^{25}_{341} + \bt^{21}_{345}
\end{align}
In Eq.~\eqref{eq:5.15}, we removed vanishing t-objects $\bt^{2341}_{45}$, $\bt^{41}_{245}$ and $\bt^{2145}_{34}$. In Eq.~\eqref{eq:5.16}, we applied atomic reduction to obtain
\[
    \bt^{23415} + \bt^{2345}_{1} = \bt^{2345}
\]
and
\[
    \bt^{415}_{2} + \bt^{45}_{21} = \bt^{45}_{2}
\]
Proceeding with this approach for the remaining multiplications, we obtain the following state vector:
\begin{align}
    & \begin{Bmatrix}
          1  &  1  &  1  &  1 &  1 &  1  & 1 & 1  \\
          0  &  1  &  1  &  1 &  1 &  0  & 1 & 1  \\
          -  &  0  &  -  &  1 &  1 &  0  & 1 & 1  \\
          0  &  1  &  0  &  0 &  1 &  0  & 1 & 1  \\
          1  &  1  &  0  &  0 &  0 &  1  & 0 & 1
    \end{Bmatrix}
\end{align}
This state vector contains only values ``1'' in the 8th column, that is, $E_8$ is identically true. This confirms that the conclusion logically follows from the assumption. Note that events $E_5$ (premise) and $E_7$ (conclusion) have identical values in all rows. Hence, a stronger statement can be made: the premise and conclusion are equivalent, expressed as $E_5 \leftrightarrow E_7$ (import-export propositional form).

This example illustrates how the process of logical inference is reduced to a sequence of simple algebraic operations on t-objects, namely, multiplication and reduction.

\section{Other Operations on State Vectors}\label{sec:other-operations-on-state-vectors}

Below, we consider two additional operations on the state vectors: index juggling and index removal. Both operations are only defined in set representations when the state vectors are considered as sets of states and state duplicates are ignored. In row decomposition, these operations result in a transformation within an equivalence class and can be used only if we are interested in the result up to equivalence transformation.

\subsection{Index Juggling}\label{subsec:index-juggling}

We define two operations: index \emph{raising}, denoted by $\bs^{:i}$, and index \emph{lowering}, denoted by $\bs_{:i}$. Index raising for event $i$ converts all holes and zero values to 1s. Conversely, index lowering replaces all holes and 1s with 0s.

Raising and lowering the index of a t-object $\bt^\alpha_\beta$ is thus defined as
\begin{align*}
    & \bt^{\alpha\ :i}_\beta = \bt^{\alpha\cup\{i\}}_{\beta \smallsetminus \{i\}} \\
    & \bt^{\alpha}_{\beta\ :i} = \bt^{\alpha\smallsetminus\{i\}}_{\beta \cup\{i\}}
\end{align*}

For an arbitrary state vector, the index is raised and lowered by raising and lowering the indices of all the t-objects it comprises. For instance:
\[
    \bs = \begin{Bmatrix}
              1 & 1 & - \\
              0 & - & 0 \\
              - & 0 & 1
    \end{Bmatrix}\,;\qquad
    \bs^{:2} = \begin{Bmatrix}
                   1 & 1 & - \\
                   0 & 1 & 0 \\
                   - & 1 & 1
    \end{Bmatrix}\,;\qquad
    \bs_{:3} = \begin{Bmatrix}
                   1 & 1 &  0 \\
                   0 & - & 0 \\
                   - & 0 & 0
    \end{Bmatrix}\,;\qquad
\]
In the row decomposition representation, the result of raising or lowering the index may differ for two vectors, even if one is an identity transformation of the other. For instance
\begin{align*}
    & \bs = \begin{Bmatrix}
                1 & - \\
                0 & 0
    \end{Bmatrix}\quad \Rightarrow\quad
    \bs^{:1} = \begin{Bmatrix}
                   1 & - \\
                   1 & 0
    \end{Bmatrix}  \\
    & \bs = \begin{Bmatrix}
                1 & 1 \\
                - & 0
    \end{Bmatrix}\quad \Rightarrow\quad
    \bs^{:1} = \begin{Bmatrix}
                   1 & 1 \\
                   1 & 0
    \end{Bmatrix}\\
\end{align*}
Therefore, index juggling can only be considered up to equivalence transformation.

For index juggling and subvector operations, the order is important if the same index is present in both subvector and juggling operations. It is easy to see that
\begin{align}
    & \bs^{n\,:n} = \bs^n\,, \quad \bs_{n\,:n} = \bs_n\,,\\
    & (\bs^{:n})^n = \bs^{:n}\,,\quad (\bs_{:n})_n = \bs_{:n}\,,\\
    & (\bs^{:n})_n = 0\,,\quad (\bs_{:n})^n = 0
\end{align}
Index juggling distributes (up to equivalence transformation)
\begin{align}
(\ba + \bb)
    ^{:n} \cong \ba^{:n} + \bb^{:n}
\end{align}
Notice that in general
\[
    (\ba\,\bb)_{:n} \not = \ba_{:n} \; \bb_{:n}\,,
\]
but
\begin{align}
(\ba^n\,\bb)
    _{:n} = (\ba^n\,\bb^n)_{:n} \cong \ba^n_{\ :n}\,\bb^n_{\ :n}
\end{align}

\subsection{Event Removal}\label{subsec:event-removal}

If event $i$ must be removed from state vector $\bs$ in matrix notation, it is equivalent to replacing all its values with holes. We designate event removal as $\bs^{-i}$ or equivalently $\bs_{-i}$ (using upper or lower index notation does not matter, and we use these interchangeably). Removing index from a t-object can be defined as
\[
    \bt^{\alpha\;-i}_\beta = \bt^{\alpha\smallsetminus\{i\}}_{\beta\smallsetminus\{i\}}
\]
Removing the index from a state vector succeeds by removing this index from all the t-objects it comprises. Fro instance
\[
    \bs = \begin{Bmatrix}
              1 & 1 & - \\
              0 & - & 0 \\
              - & 0 & 1
    \end{Bmatrix}\,;\qquad
    \bs^{-2} = \begin{Bmatrix}
                   1 & - & - \\
                   0 & - & 0 \\
                   - & - & 1
    \end{Bmatrix}
\]
Similar to index juggling, index removal can only be interpreted up to equivalence transformation because the identity transformation of an operand can produce a different outcome. For example,
\[
    (\bt^1_2)^{-3} = \bt^1_2\,,\quad\text{but}\qquad (\bt^1_2)^{-3} = (\bt^{13}_2 + \bt^1_{23})^{-3} = 2\,\bt^1_2
\]

It is easy to see that up to equivalence transformation, the event removal can be represented as
\begin{align}
    \bs^{-n} \cong \bs + (\bs_n)^{:n} + (\bs^n)_{:n}
\end{align}
Event removal distributes
\begin{align}
(\ba + \bb)
    _{-n} \cong \ba_{-n} + \bb_{-n}
\end{align}

Notice that in general case
\[
    (\bs\,\bq)_{-n} \not = \bs_{-n} \, \bq_{-n}\,,
\]
but the following holds
\begin{align}
    &(\bs\,\bq)_{n\;-n} = (\bs_n\,\bq_n)_{-n} \cong \bs_{n\;-n}\,\bq_{n\;-n}
\end{align}

\subsection{Other Inference Problems}\label{subsec:other-inference-problems}

\subsubsection{Identifying Boolean Function}

Let $\bs$ be the state vector in the row decomposition representation. We are interested in whether some event $E_n$ is a Boolean function of other events, that is, the value $e_n$ is uniquely defined by the values of the other events.

It is easy to see that the event $E_n$ is a Boolean function of other events if the state vector $\bs$ obeys
\begin{align}
    \bs_{n\;-n} \; \bs^{n\;-n} = 0
\end{align}
which implies that the rows of $\bs_n$ and $\bs^n$ are all different for the set of all events except the $n$-th. We can find an equivalent formulation using the index juggling. Substituting the definition of event removal, and noticing that $\bs_n\,\bs^n = 0$ and $\bs_n^{\ :n}\,\bs^n_{\ :n} = 0$, we obtain an equivalent pair of equations
\begin{align*}
    & \bs_n\,\bs^n_{\ :n} = 0 \\
    & \bs^n\,\bs_n^{\ :n} = 0
\end{align*}

\subsubsection{Identifying Free Events}

As discussed previously, if the state vector is not in a canonical form or is not binary, its \free columns are not necessarily trivial. Identifying \free events might be necessary, for example, in the context of higher-order logic.

Let us consider a vector $\bs$ and let $E_n$ be \free in $\bs$. This can be expressed as
\begin{align}
    \label{eq:event_independence}
    \bs_{-n} \cong \bs
\end{align}
This identity provides an algorithmic method for identifying the \free columns of a state vector.

\subsubsection{Removing Supplementary Events}\label{subsubsec:removing-supplementary-events}

Sometimes, during the inference process, it is convenient to introduce supplementary events to simplify the calculation of the state vectors corresponding to nested logical formulas. After the calculation is completed, the supplementary events can be removed from the state vector. It is important to take into account that due to
\[
    (\bs\,\bq)_{-n} \not = \bs_{-n} \, \bq_{-n}\,,
\]
event removal should be performed after the valid set is found.\footnote{It can be just a subset of the valid set, for which the event to be removed is \bound.} Indeed, suppose we have a system of two events $\{E_1, E_2\}$, and a logical formula
\[
    E_1 \vee (E_1 \wedge E_2)
\]
We introduce a supplementary event $E_3$ and rewrite the formula as a conjunction of the two formulas:
\begin{align*}
    & E_3 = E_1 \wedge E_2 \quad \sim \quad
    \bs = \begin{Bmatrix}
              1  &  1  &  1 \\
              0  &  1  &  0 \\
              1  &  0  &  0 \\
              0  &  0  &  0
    \end{Bmatrix}     \\
    & E_1 \vee E_3 \quad \sim \quad
    \bq = \begin{Bmatrix}
              1  &  -  &  1 \\
              0  &  -  &  1 \\
              1  &  -  &  0
    \end{Bmatrix}
\end{align*}
The valid set is determined as product $\bs\,\bq$. After multiplying the vectors and removing the supplementary events, we obtain the correct valid set:
\[
    (\bs\,\bq)_{-3} =
    \begin{Bmatrix}
        1  &  -
    \end{Bmatrix}
\]
If instead we removed supplementary events from the state vectors before multiplication, we would obtain the wrong result:
\[
    \bs_{-3}\,\bq_{-3} =
    \begin{Bmatrix}
        -  &  -
    \end{Bmatrix}
\]

\bibliography{bibliography}

\end{document}